\documentclass[10pt,twocolumn,letterpaper]{article}

\usepackage{cvpr}              

\usepackage{pifont}
\usepackage{amsmath,amsfonts,amssymb}
\usepackage[accsupp]{axessibility}
\usepackage[most]{tcolorbox}
\usepackage[dvipsnames]{xcolor}
\usepackage{colortbl}
\usepackage{arydshln}

\usepackage{graphicx}
\usepackage{tikz}
\usetikzlibrary{calc,positioning}

\usepackage{algorithm}
\usepackage{algpseudocode}

\usepackage{wrapfig}
\usepackage{adjustbox}

\usepackage{pifont}
\newcommand{\cmark}{\ding{51}}
\newcommand{\xmark}{\ding{55}}

\definecolor{DarkRed}{RGB}{139,0,0}  
\definecolor{DarkGreen}{RGB}{42,110,63}
\definecolor{DarkYellow}{RGB}{160, 115, 0} 

\definecolor{ada_blue}{rgb}{0,205,205}
\definecolor{glt_red}{rgb}{109,205,255}
\definecolor{MorandiBlue}{RGB}{118,134,146}
\definecolor{demphcolor}{RGB}{144,144,144}
\definecolor{mygray}{gray}{0.4}
\definecolor{autopurple}{HTML}{7030A0}
\definecolor{dyna_yellow}{HTML}{BF9000}
\definecolor{adaptive_blue}{HTML}{0070C0}
\definecolor{darkgrey}{RGB}{120,120,120}
\definecolor{mygrey}{RGB}{200,200,200}
\definecolor{myblue}{HTML}{00CDCD}
\definecolor{champagne}{rgb}{0.97, 0.91, 0.81}
\definecolor{darksalmon}{rgb}{0.91, 0.59, 0.48}
\definecolor{emerald}{rgb}{0.31, 0.78, 0.47}
\definecolor{green(pigment)}{rgb}{0.0, 0.65, 0.31}
\definecolor{amaranth}{rgb}{0.9, 0.17, 0.31}
\definecolor{iris}{rgb}{0.35, 0.31, 0.81}
\definecolor{uu}{rgb}{0.95, 0.51, 0.51}
\definecolor{spirodiscoball}{rgb}{0.06, 0.75, 0.99}
\definecolor{cadetblue}{RGB}{95,158,160} 
\definecolor{keywordcolor}{RGB}{178,34,34} 

\definecolor{customgreen}{HTML}{667b5b}

\definecolor{customblue}{HTML}{bcccea}

\colorlet{mag_fg}{amaranth}
\colorlet{dir_fg}{adaptive_blue}
\definecolor{dist_deep}{RGB}{52,97,57}
\colorlet{dist_fg}{dist_deep}
\definecolor{mag_bg}{RGB}{255,235,238}   
\definecolor{dir_bg}{RGB}{232,242,255}   
\definecolor{dist_bg}{RGB}{232,245,233}  

\colorlet{magfg}{mag_fg}
\colorlet{dirfg}{dir_fg}
\colorlet{distfg}{dist_fg}
\colorlet{magbg}{mag_bg}
\colorlet{dirbg}{dir_bg}
\colorlet{distbg}{dist_bg}

\newcommand{\blueupsmall}[1]{$_{\color{RoyalBlue}\uparrow \text{#1}}$}
\newcommand{\reddownsmall}[1]{$_{\color{OrangeRed}\downarrow \text{#1}}$}

\newcommand{\blueup}[1]{\textcolor{RoyalBlue}{$\uparrow$ #1}}
\newcommand{\reddown}[1]{\textcolor{OrangeRed}{$\downarrow$ #1}}

\makeatletter
\newcommand{\thickhline}{%
  \noalign{\ifnum0=`}\fi\hrule height 1pt\relax
  \futurelet\reserved@a\@xhline
}
\makeatother

\tcbset{highlight math/.append style={left=0mm,right=0mm,top=0mm,bottom=0mm, colframe=white}}

\usepackage[table]{xcolor}
\usepackage{multirow}

\newcommand{\rowstyle}[1]{\gdef\currentrowstyle{#1}\leavevmode#1\ignorespaces}
\newcommand{\tblgray}{gray!95}

\newcommand{\grayrow}{\rowstyle{\color{\tblgray}}}

\definecolor{cvprblue}{rgb}{0.21,0.49,0.74}
\usepackage[pagebackref,breaklinks,colorlinks,citecolor=cvprblue,urlcolor=cvprblue,linkcolor=cvprblue]{hyperref}
\def\paperID{***} 
\def\confName{CVPR}
\def\confYear{2026}

\title{Fine-Tuning Impairs the Balancedness of Foundation Models in  \\
Long-tailed Personalized Federated Learning}

\author{
    Shihao Hou$^{1}$\hspace{20pt}
    Chikai Shang$^{1}$\hspace{20pt}
    Zhiheng Yang$^{1}$\hspace{20pt}
    Jiacheng Yang$^{1}$\hspace{20pt}
    Xinyi Shang$^{2}$ \\
    Junlong Gao$^{1}$\hspace{20pt}
    Yiqun Zhang$^{3}$\hspace{20pt}
    Yang Lu$^{1}$\thanks{Corresponding author: Yang Lu (luyang@xmu.edu.cn).\protect\\\protect\hspace*{1.8em}Accepted by CVPR 2026.} \\
    {\footnotesize $^1$Key Laboratory of Multimedia Trusted Perception and Efficient Computing, Ministry of Education of China, Xiamen University, Xiamen, China} \\
    {\footnotesize $^2$Department of Statistical Science, University College London, London, United Kingdom} \\
    {\footnotesize $^3$School of Computer Science and Technology, Guangdong University of Technology, Guangzhou, China} \\
    {\tt\small houshihao@stu.xmu.edu.cn\hspace{20pt}luyang@xmu.edu.cn}
}

\begin{document}

\maketitle

\begin{abstract}

\vspace{-0.5em}
Personalized federated learning (PFL) with foundation models has emerged as a promising paradigm enabling clients to adapt to heterogeneous data distributions. However, real-world scenarios often face the co-occurrence of non-IID data and long-tailed class distributions, presenting unique challenges that remain underexplored in PFL. In this paper, we investigate this long-tailed personalized federated learning and observe that current methods suffer from two limitations: (i) Fine-tuning degrades performance below zero-shot baselines due to the erosion of inherent class balance in foundation models; (ii) Conventional personalization techniques further transfer this bias to local models through parameter or feature-level fusion. To address these challenges, we propose \textbf{Fed}erated Learning via Gradient \textbf{Pu}rification and \textbf{Re}sidual \textbf{L}earning (FedPuReL), which preserves balanced knowledge in the global model while enabling unbiased personalization. Specifically, we purify local gradients using zero-shot predictions to maintain a class-balanced global model, and model personalization as residual corrections atop the frozen global model. Extensive experiments demonstrate that FedPuReL consistently outperforms state-of-the-art methods, achieving superior performance on both global and personalized models across diverse long-tailed scenarios. The code is available at \url{https://github.com/shihaohou/FedPuReL}.
\vspace{-1em}

\end{abstract}

\vspace{-1em}
\section{Introduction}

Personalized federated learning (PFL) has emerged as a promising solution to address data heterogeneity in collaborative learning while preserving privacy. Recently, with the rise of pre-trained foundation models such as CLIP~\cite{clip}, parameter-efficient fine-tuning (PEFT) has made it viable to adapt these powerful models in federated settings, which updates a limited number of parameters while keeping the remaining ones frozen~\cite{yangfeddda, bai2024diprompt, li2021model}. Recent work~\cite{guo2023pfedprompt, li2024fedotp, cui2024fedpgp, pan2024promptfolio} has demonstrated the effectiveness of this paradigm through personalized parameter mechanisms that construct separate global and local trainable parameters, leveraging aggregated global knowledge as a bridge to enhance client-specific adaptation.

\begin{figure}[!t]
\centering
 \includegraphics[width=0.49\textwidth]{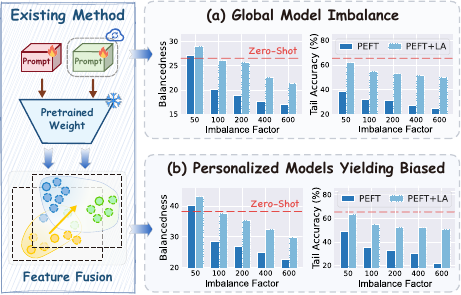}
\vspace{-1.4em}
\caption{\textbf{Imbalanced Global Model Compromises Personalization.} We use the \textit{Balancedness metric}~\cite{kang2020exploring} to measure performance distribution across classes (detailed in Sec.~\ref{subsec:motivation}). Under the Fed-LT scenario, we observe that \textbf{(a)} aggregating global parameters damages the foundation model's balanced knowledge, degrading below zero-shot baselines, while \textbf{(b)} feature fusion transfers this imbalance to personalized models, yielding similarly biased predictions. Even Logit Adjustment (LA)~\cite{menon2021la}, which leverages class priors, fails to restore zero-shot balancedness.}
 \label{fig:motivation1}
 \vspace{-10pt}
\end{figure}

However, existing PEFT-based PFL methods have primarily focused on statistical heterogeneity under the assumption of balanced data distributions. In reality, federated learning scenarios often encounter non-IID data and long-tailed class distributions, termed federated long-tailed learning (Fed-LT)~\cite{li2024federated, xiao2024fedloge, zhang2024rucr}, where scarce tail-class samples are dispersed across heterogeneous clients. This co-occurrence causes aggregated global models to become heavily biased toward head classes, hindering effective learning of tail-class representations~\cite{shang2022federated, hou2025capt}. While current PEFT-based PFL methods~\cite{guo2023pfedprompt, li2024fedotp} demonstrate effectiveness in handling heterogeneity, they fail to preserve the balanced knowledge inherent in pre-trained foundation models when facing such long-tailed distributions. 

Conversely, existing Fed-LT methods~\cite{sarkar2020fed, wang2021ratioloss, hou2025capt, shi2024clip2fl} primarily focus on learning a single balanced global model through techniques such as loss re-balancing or knowledge distillation, but overlook client-specific variations. Notably, local data distributions can diverge significantly from the global long-tailed pattern, with clients potentially having different tail classes or varying degrees of imbalance.

Motivated by these limitations, our work investigates how to better leverage foundation models for long-tailed personalized federated learning. Through empirical analysis, as illustrated in Fig.~\ref{fig:motivation1}, we identify two critical challenges that recent approaches fail to address. For the global model, naive fine-tuning significantly degrades its balancedness, leaving it \textit{even less balanced than the original zero-shot model} in severely long-tailed scenarios. This raises our first question: \hypertarget{q1}{\textbf{I)} \textbf{\textit{Balancedness: How can we learn more balanced global models in Fed-LT by leveraging rather than losing the original balanced knowledge of foundation models?}}} For personalized models, existing methods achieve personalization through parameter or feature-level fusion, which inevitably propagates the imbalance of the global to personalized models, yielding similarly biased predictions. Thus, this motivates our second question: \hypertarget{q2}{\textbf{II)}} \textbf{\textit{Personalization: How can we enable effective client personalization without inheriting global bias?}}

To address these questions, we develop two key insights. For \hyperlink{q1}{I)}, we leverage the balanced characteristics of zero-shot CLIP predictions to constrain local updates, regularizing them toward the pre-trained model's equilibrium. This enables clients to adapt to downstream tasks without introducing head-class bias. In response to \hyperlink{q2}{II)}, we reconceptualize personalization as residual learning in the output space, where client-specific adaptations become additive logit corrections applied to a frozen global model. This decouples personalization from global parameters, allowing local residual branches to capture client-specific deviations while preserving the global model's balanced representations. Building on these insights, we propose \textbf{Fed}erated Learning via Gradient \textbf{Pu}rification and \textbf{Re}sidual \textbf{L}earning (FedPuReL), which preserves balanced knowledge in the global model while enabling unbiased personalization. Our principal contributions are summarized as follows:

\begin{itemize}[leftmargin=*]
    \item[\ding{182}] We reveal that fine-tuning in Fed-LT inevitably erodes their inherent balanced knowledge, leading to significant bias and diminished tail-class performance.

    \item[\ding{183}] We propose FedPuReL, a balance-preserving personalization framework that constrains local gradients via zero-shot guidance to maintain a class-balanced global model while enabling residual learning for unbiased client-specific adaptations.
    
    \item[\ding{184}] We demonstrate superior and consistent performance improvements over strong baselines across multiple datasets with varying imbalance factors, surpassing even long-tailed strategies that leverage class priors.
\end{itemize}

\vspace{-0.5em}
\section{Related Work}

\subsection{Federated Long-Tailed Learning}

Federated long-tailed learning addresses class imbalance in FL environments, where the global long-tailed distribution intensifies statistical heterogeneity and biases global models toward head classes~\cite{li2020federated,zhang2024long, luo2025long, lt2026hong, Shang_2025_ICCV}. Recent studies have mainly pursued the following directions: (1) Loss-based rebalancing methods work by modifying loss functions to down-weight frequent classes~\cite{sarkar2020fed,wang2021ratioloss,Yan_2025_ICCV}; (2) Classifier decoupling approaches separate representation learning from classifier training to achieve balanced optimization on the server~\cite{shang2022federated}; (3) Gradient-based balancing techniques employ self-adjusting mechanisms or client grouping strategies to achieve class-wise equilibrium through accumulated gradient analysis~\cite{Fed-Grab,zeng2023gbme}. However, these methods focus solely on learning a single balanced global model, overlooking the diverse needs of individual clients. In practice, local data distributions can diverge significantly from the global long-tailed pattern. Clients may have different tail classes or varying imbalance degrees, necessitating personalized adaptations beyond the global model.

\subsection{Personalized Federated Learning}

Personalized federated learning extends traditional FL by developing client-specific models that adapt to local data distributions while benefiting from collaborative global knowledge~\cite{tan2022towards, kairouz2021advances, marfoq2021federated}. This approach effectively addresses data heterogeneity through various strategies: regularization-based methods balance local adaptation with global consistency via proximal terms~\cite{li2021ditto,dinh2020pfedme}, while model decomposition approaches separate shared representations from personalized components~\cite{arivazhagan2019fedper,deng2020apfl,shamsian2021pfedhn}. However, these methods implicitly assume balanced datasets and fail to address the compounded challenges when both non-IID and class imbalance coexist. While recent work like FedLoGe~\cite{xiao2024fedloge} attempts to integrate personalization with long-tailed learning through neural collapse, existing approaches have yet to leverage the balanced knowledge inherent in foundation models to tackle this dual challenge.

\subsection{Federated Learning with Foundation models}

Leveraging foundation models in federated learning has emerged as a promising direction to enhance performance and efficiency by fine-tuning large pre-trained models rather than training from scratch. This paradigm shift addresses communication bottlenecks through parameter-efficient fine-tuning (PEFT) techniques that transmit and update only small parameter subsets. Prompt-based methods represent a major thread in this area~\cite{guo2023promptfl}, enabling clients to learn lightweight prompt parameters with subsequent work incorporating personalization strategies~\cite{pan2024promptfolio,guo2023pfedprompt} and other techniques~\cite{li2024fedotp,cui2024fedpgp,iverson2024pfedmoap} to handle client heterogeneity. Adapter and low-rank adaptation strategies offer complementary approaches through fine-tuning lightweight adapter modules or selectively aggregating low-rank matrices~\cite{lu2023fedclip, wang2024fedsalora}. Despite these advances in addressing data heterogeneity and parameter efficiency, existing foundation-model-based FL methods do not explicitly tackle class imbalance arising from long-tailed federated distributions. This gap motivates our work to integrate the representational strengths of pre-trained models with targeted mechanisms for long-tail balancing and personalization.

\section{Methodology}

\subsection{Preliminaries}

\noindent\textbf{Federated Learning with Long-Tailed Data.} We consider a FL framework with $K$ clients coordinated by a central server. Each client $k \!\in\! \{1, 2, \ldots, K\}$ holds a local dataset $\mathcal{D}_k \!=\! \{(\mathbf{x}_i, y_i)\}_{i=1}^{n_k}$. At each communication round $t$, a subset $\mathcal{S}_t \!\subseteq\! \{1, 2, \ldots, K\}$ of clients is selected to participate in training. For non-IID distributions, heterogeneity among clients is modeled using a Dirichlet distribution, where a smaller $\alpha_{\text{dir}}$ indicates higher client heterogeneity, demonstrating greater divergence between local and global distributions. For long-tailed distributions, the global dataset $\mathcal{D} \!=\! \bigcup_{k=1}^K \!\mathcal{D}_k$ follows a class-imbalanced distribution with class frequencies sorted as $n_1 \!\geq\! n_2 \!\geq\! \dots \!\geq\! n_C$, where $n_c \!=\! \sum_{k=1}^{K} \!n_c^k$ denotes the total number of samples in class $c$. The degree of imbalance is defined as $\mathrm{IF} \!=\! n_1 / n_C$.

\noindent\textbf{CLIP and Parameter-Efficient Fine-Tuning.} We build upon CLIP~\cite{clip}, which performs zero-shot classification by computing cosine similarity between image and text embeddings of class prompts. To adapt CLIP to downstream tasks while preserving its zero-shot generalization, we employ PEFT methods that introduce learnable parameters $\boldsymbol{\phi}$ while freezing the pre-trained parameters $\boldsymbol{W}$.

\begin{figure}[t]
\centering
\begin{tikzpicture}
  \node[inner sep=0, outer sep=0] (img) {\includegraphics[width=\linewidth]{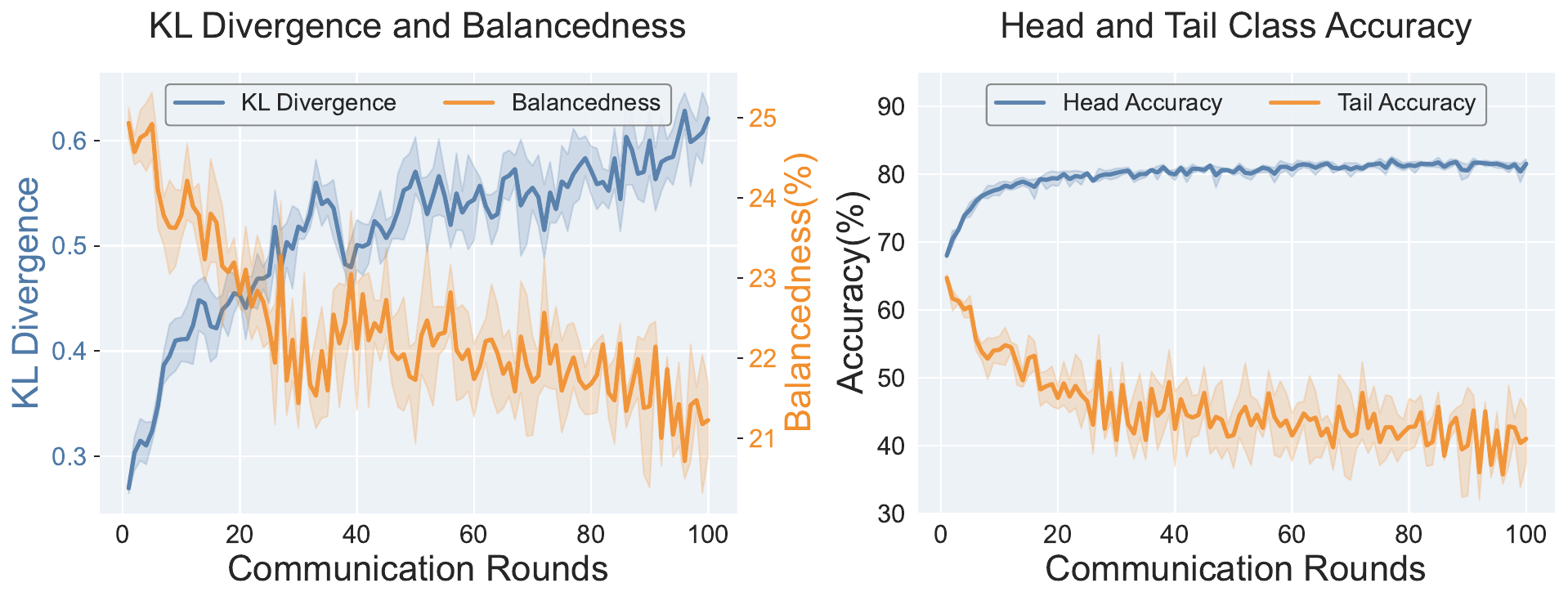}};
  \vspace{-30pt}
  \node[anchor=north, font=\footnotesize, yshift=-0.5ex]
  at ($(img.south west)!0.25!(img.south east)$) {\footnotesize(a)};
\node[anchor=north, font=\footnotesize, yshift=-0.5ex]
  at ($(img.south west)!0.75!(img.south east)$) {\footnotesize(b)};
\end{tikzpicture}
\vspace{-22pt}
\caption{\textbf{Balance degradation during federated fine-tuning.} \textbf{(a)} TKL divergence and balancedness exhibit strong negative correlation, indicating that diverging from zero-shot predictions degrades model balance. \textbf{(b)} Head-class accuracy increases while tail-class accuracy degrades, revealing amplified head-class bias.}
\vspace{-1em}
\label{fig:motivation2}
\end{figure}

\subsection{A Closer Look at the Zero-Shot Model}
\label{subsec:motivation}

To investigate the role of zero-shot foundation models in federated long-tailed learning, we conduct a preliminary analysis examining how fine-tuning affects the model's predictions and class balance. We employ two key metrics to characterize this evolution.

\begin{figure*}[!t]
\centering
 \includegraphics[width=0.99\textwidth]{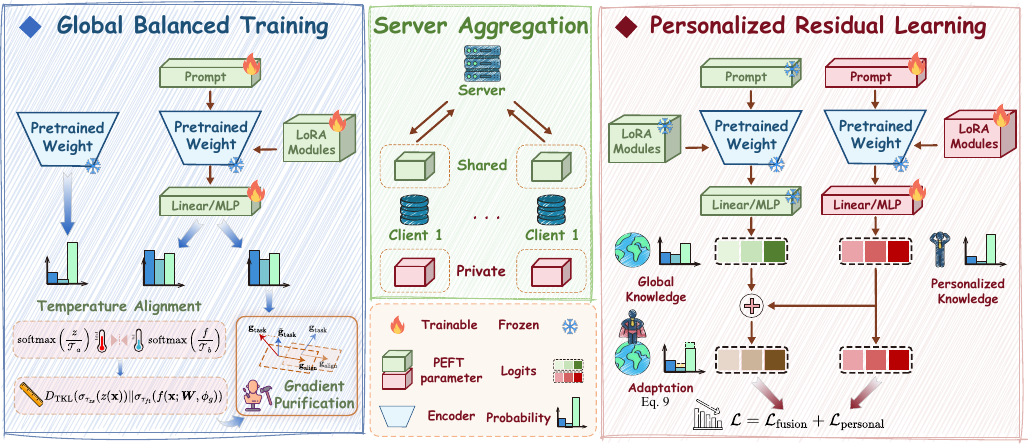}
 \caption{\textbf{Overview of FedPuReL framework.} (Left) \textit{Global Balanced Training:} Clients train shared PEFT parameters (e.g., LoRA modules) using zero-shot guided gradient alignment. Temperature-aligned predictions enable gradient purification via projection. (Center) \textit{Server Aggregation:} Shared parameters are aggregated; personalized parameters remain private. (Right) \textit{Personalized Residual Learning:} Personalization stage learns client-specific residuals over the frozen global parameters through dual-branch additive fusion.}
 \label{fig:method}
 \vspace{-12pt}
\end{figure*}

\noindent\textbf{Temperature-Aligned KL Divergence.} 
We introduce temperature-aligned KL divergence to quantify the structural distributional shift between fine-tuned and zero-shot predictions while neutralizing differences in prediction confidence. Standard KL divergence captures all distributional deviations without distinguishing their nature; for instance, fine-tuning on balanced data also shifts predictions, but this shift reflects improved confidence rather than bias. Our temperature-aligned metric normalizes logit scales to isolate structural bias across classes.

For a sample $\mathbf{x}$ with fine-tuned logits $\mathbf{l} \in \mathbb{R}^C$ and zero-shot logits $\mathbf{z} \in \mathbb{R}^C$, we define the temperature-softmax as $\sigma_{\tau}(\mathbf{a})=\operatorname{softmax}(\mathbf{a}/\tau)$ and the entropy function as $H(\mathbf{p})=-\sum_{c=1}^{C} p_c \log p_c$. We align both distributions to a common target entropy $H^\star = \tfrac{1}{2}\big[ H(\sigma_{1}(\mathbf{l})) + H(\sigma_{1}(\mathbf{z})) \big]$ by computing the aligned temperatures:
\begin{equation}
\tau_{a} = H_{a}^{-1}(H^\star), \quad \text{where} \quad H_{a}(\tau) = H(\sigma_\tau(\mathbf{a})).
\end{equation}
The temperature-aligned KL divergence is then defined as:
\begin{equation}
\begin{aligned}
\mathrm{TKL}(\mathbf{x})
&= D_{\mathrm{KL}}\!\left(\sigma_{\tau_{f}}(\mathbf{f}) \,\middle\Vert\, \sigma_{\tau_{z}}(\mathbf{z})\right) \\
&= \sum_{c=1}^{C} \sigma_{\tau_{f}}(\mathbf{f})_c
   \log \frac{\sigma_{\tau_{f}}(\mathbf{l})_c}{\sigma_{\tau_{z}}(\mathbf{z})_c}.
\end{aligned}
\end{equation}
This metric neutralizes global confidence differences, preserving focus on class bias.

\noindent\textbf{Balancedness Metric.}
We adopt a balancedness metric, introduced in~\cite{kang2020exploring}, aiming to measure how balanced the model is by computing performance equity across classes:
\begin{equation}
    \beta(V) = \frac{1}{C^2} \sum_{i,j} \exp\left(-\frac{|a_i - a_j|^2}{\sigma}\right),
\end{equation}
where $a_i$ denotes the accuracy of class $i$ and $\sigma$ is a fixed scaling parameter. This metric reaches its maximum value of 1 when all classes achieve equal accuracy, directly quantifying cross-class performance equity.

Using these two metrics, we analyze the fine-tuning dynamics of CLIP on federated long-tailed scenarios. Specifically, we track both TKL divergence and balancedness $\beta(V)$ across training rounds to examine how fine-tuning affects the model's predictions and class balance. Our analysis reveals two critical patterns:

\noindent\textbf{Observation 1: Negative correlation between divergence and balance.} As shown in Fig.~\ref{fig:motivation2}(a), TKL divergence and balancedness $\beta(V)$ exhibit a strong negative correlation across training rounds. As fine-tuning progresses, predictions increasingly deviate from the zero-shot model, while class balance concurrently deteriorates. This reveals that adapting to long-tailed downstream tasks gradually erodes the balanced knowledge inherent in the pre-trained model.

\noindent\textbf{Observation 2: Amplified head-class bias.}
Fig.~\ref{fig:motivation2}(b) shows that fine-tuning disproportionately benefits head classes while degrading tail-class performance, significantly widening the performance gap. This asymmetric adaptation actively amplifies distributional bias, directly contributing to the declining balancedness.

These observations reveal that fine-tuning foundation models on long-tailed data creates a fundamental trade-off between task adaptation and preserving the balanced knowledge inherent in zero-shot predictions. Pre-trained CLIP models possess robust cross-class equilibrium from large-scale diverse training, but this equilibrium deteriorates as they adapt to imbalanced downstream distributions.

\subsection{Proposed Method}

We propose FedPuReL to learn PEFT parameters that adapt foundation models to downstream tasks while preserving their inherent balanced representations, as depicted in Fig.~\ref{fig:method}. We employ gradient purification to maintain global balance during collaborative training, and residual learning to achieve unbiased client-specific adaptation. This ensures task adaptation without amplifying class imbalances.

\subsubsection{Balancedness: Gradient Purification }

As demonstrated in Sec.~\ref{subsec:motivation}, naive fine-tuning inevitably shifts the model away from balanced zero-shot representations, causing performance degradation on tail classes. To address this, we constrain local gradient updates to directions that do not deviate from the zero-shot model, thereby preserving its balanced knowledge while adapting to downstream tasks. Specifically, we encourage the fine-tuned model to remain aligned with the balanced distribution of the zero-shot model through TKL divergence:
\begin{equation}
\mathcal{L}_{\text{align}} = D_{\text{TKL}} \left( \sigma_{\tau_{zs}} (\mathbf{z}(\mathbf{x})) \,\Big\|\, \sigma_{\tau_{ft}} (\mathbf{f}(\mathbf{x}; \boldsymbol{W}, \boldsymbol{\phi}_g)) \right),
\end{equation}
where $\mathbf{z}$ and $\mathbf{f}$ denote the zero-shot and fine-tuned logits, respectively, and $\tau_{zs}$, $\tau_{ft}$ are temperature parameters controlling the smoothness of the softmax distributions. The gradient $\mathbf{g}_{\text{align}} = \nabla_{\boldsymbol{\phi}_g} \mathcal{L}_{\text{align}}$ points toward preserving the balanced knowledge encoded in the zero-shot model.

To preserve this balanced knowledge, we project the gradient onto the subspace orthogonal to $\mathbf{g}_{\text{align}}$:
\begin{equation}
\tilde{\mathbf{g}}_{\text{task}} = \begin{cases}
    \mathbf{g}_{\text{task}}, & \text{if } \langle\mathbf{g}_{\text{task}}, \mathbf{g}_{\text{align}}\rangle \geq 0 \\[6pt]
    \mathbf{g}_{\text{task}} - \dfrac{\langle\mathbf{g}_{\text{task}}, \mathbf{g}_{\text{align}}\rangle}{\|\mathbf{g}_{\text{align}}\|^2} \mathbf{g}_{\text{align}}, & \text{otherwise.}
\end{cases}
\end{equation}
where $\mathbf{g}_{\text{task}} = \nabla_{\boldsymbol{\phi}_g} \text{CE}(y, \sigma(\mathbf{f}(\mathbf{x}; \boldsymbol{W}, \boldsymbol{\phi}_g)))$. When the task gradient conflicts with $\mathbf{g}_{\text{align}}$, naively applying it would push the model away from the pre-trained anchor. This projection removes any component of $\mathbf{g}_{\text{task}}$ that opposes $\mathbf{g}_{\text{align}}$, ensuring that every update preserves balanced knowledge while adapting to downstream tasks.

After local training with the purified gradient, clients upload parameters to the server for aggregation:
\begin{equation}
\boldsymbol{\phi}_g^{(t+1)} \leftarrow \sum_{k \in \mathcal{S}_t} \frac{n_k}{\sum_{j \in \mathcal{S}_t} n_j} \boldsymbol{\phi}_g^{k,(t)},
\end{equation}
where $\boldsymbol{\phi}_g^{k,(t)}$ is obtained through local updates using $\tilde{\mathbf{g}}_{\text{task}}$. The resulting global model $(\boldsymbol{W}, \boldsymbol{\phi}_g^{(t+1)})$ achieves balanced, task-adapted representations across clients, providing a stable foundation for client-specific personalization.

\subsubsection{Personalization: Learning Clients Residuals}
\label{subsec:personalization}

To achieve personalization, recent methods fuse global and local information at the feature or parameter level to construct personalized models~\cite{li2021ditto, li2024fedotp, pan2024promptfolio}. However, such fusion strategies inevitably transfer class imbalances to local models, which limits their personalization effectiveness. Instead, we model personalization as learning corrective residuals that build upon the balanced global foundation. Specifically, these residuals capture client-specific patterns as additive corrections atop the global model's balanced, task-adapted knowledge. We realize this through a dual-branch architecture with two parallel pathways:

\noindent\textbf{Global Branch.} This branch produces baseline predictions:
\begin{equation}
\mathbf{l}_{\text{G}}(\mathbf{x}) = f(\mathbf{x}; \boldsymbol{W}, \boldsymbol{\phi}_g),
\end{equation}
where $f(\cdot)$ denotes the forward pass through the global model. Critically, $\boldsymbol{\phi}_g$ remains frozen throughout the personalization stage, serving as an immutable anchor that preserves the balanced, task-adapted knowledge.

\noindent\textbf{Personalized Branch.} In parallel, each client $k$ learns client-specific parameters $\boldsymbol{\phi}_k$ that capture local patterns:
\begin{equation}
\mathbf{l}_{\text{P}}^k(\mathbf{x}) = f(\mathbf{x}; \boldsymbol{W}, \boldsymbol{\phi}_k).
\end{equation}
The personalized parameters $\boldsymbol{\psi}_k$ can be flexibly instantiated as an independent classification head, additional adapter layers, learnable prompts, or LoRA modules.

\noindent\textbf{Additive Fusion.} The final prediction combines both branches additively:
\begin{equation}
\mathbf{l}_{\text{final}}^k(\mathbf{x}) =
\underbrace{\tcbhighmath[colback=distbg]{f(\mathbf{x}; \boldsymbol{W}, \boldsymbol{\phi}_g)}}_{\text{\color{distfg}{\textbf{Global Part}}}}
\;+\;
\underbrace{\tcbhighmath[colback=magbg]{f(\mathbf{x}; \boldsymbol{W}, \boldsymbol{\phi}_k)}}_{\text{\color{magfg}{\textbf{Personal Part}}}}.
\end{equation}

The personalized model inherently stands on the shoulders of giants, benefiting from the global model's balanced knowledge while adapting to local specifics. Even under local overfitting or shifts, the global branch offers a reliable anchor, averting severe degradation.

During local training, we optimize a composite loss:
\begin{align}
\mathcal{L}_{\text{fusion}}^k   &= \operatorname{CE}\!\bigl(y,\,
  \sigma(\mathbf{f}_{\mathrm{G}}(\mathbf{x})+\mathbf{f}_{\mathrm{P}}^k(\mathbf{x}))\bigr),\\
\mathcal{L}_{\text{personal}}^k &= \operatorname{CE}\!\bigl(y,\,
  \sigma(\mathbf{f}_{\mathrm{P}}^k(\mathbf{x}))\bigr),\\
\mathcal{L}_{\text{total}}^k    &= (1-\lambda)\,\mathcal{L}_{\text{fusion}}^k
  + \lambda\,\mathcal{L}_{\text{personal}}^k.
  \label{fusion loss}
\end{align}
where $\lambda$ controls the trade-off. The fusion loss $\mathcal{L}_{\text{fusion}}^k$ maintains stability by optimizing the combined prediction, while the personalization loss $\mathcal{L}_{\text{personal}}^k$ drives parameter specialization to capture local patterns. Gradients flow solely through $\boldsymbol{\phi}_k$, safeguarding global balance.

During inference, global model only uses the global branch $p(y|\mathbf{x}) = \sigma(f(\mathbf{x}; \boldsymbol{W}, \boldsymbol{\phi}_g)).$ For personalized inference, we fuse both branches: $p_k(y|\mathbf{x}) = \sigma(f(\mathbf{x}; \boldsymbol{W}, \boldsymbol{\phi}_g) + f(\mathbf{x}; \boldsymbol{W}, \boldsymbol{\phi}_k)).$ This fusion combines the robust global model with client-specific corrections, achieving superior accuracy on local distributions while maintaining the stability inherited from balanced global training.

\begin{table*}[t]
\centering
\caption{\textbf{Comparison with state-of-the-art methods on ImageNet-LT and Places-LT} for global models (GM) and personalized models (PM) across all classes and shot categories (Many, Med., Few). Best in \textbf{bold}, second-best \underline{underlined}.}
\footnotesize
\setlength\tabcolsep{2pt}
\renewcommand\arraystretch{1.1}
\begin{tabular}{r||cccc|cccc|cccc|cccc}
\hline\thickhline
\rowcolor{gray!20}
\multirow{3}{*}{\textbf{Method}} & \multicolumn{8}{c|}{\textbf{ImageNet-LT}} & \multicolumn{8}{c}{\textbf{Places-LT}} \\
\cline{2-17}
\rowcolor{gray!20}
 & \multicolumn{4}{c}{\textbf{GM}} & \multicolumn{4}{c|}{\textbf{PM}} & \multicolumn{4}{c}{\textbf{GM}} & \multicolumn{4}{c}{\textbf{PM}} \\
\cline{2-17}
\rowcolor{gray!20}
\multirow{-3}{*}{\textbf{Method}} & \textbf{All} & \textbf{Many} & \textbf{Med.} & \textbf{Few} & \textbf{All} & \textbf{Many} & \textbf{Med.} & \textbf{Few} & \textbf{All} & \textbf{Many} & \textbf{Med.} & \textbf{Few} & \textbf{All} & \textbf{Many} & \textbf{Med.} & \textbf{Few} \\
\hline\hline
Zero-shot &  67.05 & 67.88 & 66.49 & 66.65 & 66.68 & 67.71 & 66.07 & 65.97 & 35.55 & 33.39 & 35.21 & 40.31 & 35.33 & 33.59 & 35.21 & 40.63 \\
\hdashline

\multicolumn{17}{l}{\textcolor{gray!60}{\textit{Prompt-based}}} \\
\cellcolor{gray!10}PromptFL~\cite{guo2023promptfl}
 & \cellcolor{gray!10}67.76 & \cellcolor{gray!10}76.20 & \cellcolor{gray!10}63.71 & \cellcolor{gray!10}\underline{58.36}
 & \cellcolor{gray!10}67.51 & \cellcolor{gray!10}75.48 & \cellcolor{gray!10}63.69 & \cellcolor{gray!10}\underline{58.82}
 & \cellcolor{gray!10}38.90 & \cellcolor{gray!10}51.07 & \cellcolor{gray!10}32.86 & \cellcolor{gray!10}29.64
 & \cellcolor{gray!10}36.49 & \cellcolor{gray!10}48.00 & \cellcolor{gray!10}31.37 & \cellcolor{gray!10}27.42 \\

PromptFolio~\cite{pan2024promptfolio}
 & \underline{69.64} & \textbf{80.26} & \underline{66.37} & 47.02
 & 67.65 & \underline{77.23} & 64.89 & 45.99
 & 40.55 & 52.57 & 36.99 & \underline{31.58}
 & 37.95 & 47.73 & 34.52 & 28.32 \\

\cellcolor{gray!10}FedOTP~\cite{li2024fedotp}
 & \cellcolor{gray!10}57.03 & \cellcolor{gray!10}69.57 & \cellcolor{gray!10}59.58 & \cellcolor{gray!10}43.74
 & \cellcolor{gray!10}50.49 & \cellcolor{gray!10}66.32 & \cellcolor{gray!10}42.60 & \cellcolor{gray!10}34.14
 & \cellcolor{gray!10}36.07 & \cellcolor{gray!10}48.18 & \cellcolor{gray!10}32.23 & \cellcolor{gray!10}22.66
 & \cellcolor{gray!10}32.56 & \cellcolor{gray!10}43.28 & \cellcolor{gray!10}29.26 & \cellcolor{gray!10}21.11 \\

FedPGP~\cite{cui2024fedpgp}
 & 67.05 & 76.99 & 62.99 & 53.60
 & 66.44 & 76.59 & 62.32 & 52.97
 & 38.92 & 50.82 & 32.89 & 30.27
 & 37.36 & \underline{48.21} & 32.04 & 28.90 \\

\cellcolor{gray!10}
PromptFolio+Fed-Grab~\cite{Fed-Grab}
 & \cellcolor{gray!10}69.53 & \cellcolor{gray!10}\underline{79.90} & \cellcolor{gray!10}66.16 & \cellcolor{gray!10}52.83
 & \cellcolor{gray!10}\underline{68.14} & \cellcolor{gray!10}\textbf{77.32} & \cellcolor{gray!10}\underline{65.09} & \cellcolor{gray!10}52.11
 & \cellcolor{gray!10}\underline{41.99} & \cellcolor{gray!10}\underline{52.82} & \cellcolor{gray!10}\underline{38.43} & \cellcolor{gray!10}30.82
 & \cellcolor{gray!10}\underline{39.14} & \cellcolor{gray!10}47.82 & \cellcolor{gray!10}\textbf{36.04} & \cellcolor{gray!10}\underline{30.76} \\

\cellcolor[HTML]{D7F6FF}\textbf{FedPuReL (ours)}
 & \cellcolor[HTML]{D7F6FF}\textbf{72.96} & \cellcolor[HTML]{D7F6FF}77.32 & \cellcolor[HTML]{D7F6FF}\textbf{68.41} & \cellcolor[HTML]{D7F6FF}\textbf{66.70}
 & \cellcolor[HTML]{D7F6FF}\textbf{70.12} & \cellcolor[HTML]{D7F6FF}76.34 & \cellcolor[HTML]{D7F6FF}\textbf{67.68} & \cellcolor[HTML]{D7F6FF}\textbf{66.62}
 & \cellcolor[HTML]{D7F6FF}\textbf{43.88} & \cellcolor[HTML]{D7F6FF}\textbf{55.44} & \cellcolor[HTML]{D7F6FF}\textbf{40.94} & \cellcolor[HTML]{D7F6FF}\textbf{39.06}
 & \cellcolor[HTML]{D7F6FF}\textbf{39.68} & \cellcolor[HTML]{D7F6FF}\textbf{49.85} & \cellcolor[HTML]{D7F6FF}\underline{35.55} & \cellcolor[HTML]{D7F6FF}\textbf{32.72} \\

\textcolor{gray!60}&
\blueup{3.32} & \reddown{2.94} & \blueup{2.04} & \blueup{8.34} &
\blueup{1.98} & \reddown{0.98} & \blueup{2.59} & \blueup{7.80} &
\blueup{1.89} & \blueup{2.62} & \blueup{2.51} & \blueup{7.48} &
\blueup{0.54} & \blueup{1.64} & \reddown{0.49} & \blueup{1.96} \\

\hline
\multicolumn{17}{l}{\textcolor{gray!60}{\textit{LoRA-based}}} \\
\cellcolor{gray!10}
CLIPLoRA~\cite{cliplora}
 & \cellcolor{gray!10}70.18 & \cellcolor{gray!10}\textbf{80.91} & \cellcolor{gray!10}65.84 & \cellcolor{gray!10}52.90
 & \cellcolor{gray!10}66.07 & \cellcolor{gray!10}\textbf{77.82} & \cellcolor{gray!10}61.77 & \cellcolor{gray!10}48.86
 & \cellcolor{gray!10}41.61 & \cellcolor{gray!10}\underline{54.51} & \cellcolor{gray!10}36.49 & \cellcolor{gray!10}29.79
 & \cellcolor{gray!10}37.21 & \cellcolor{gray!10}49.10 & \cellcolor{gray!10}33.14 & \cellcolor{gray!10}25.30 \\

FedSA-LoRA~\cite{wang2024fedsalora}
 & 69.44 & \underline{80.07} & \underline{66.84} & \underline{58.83}
 & 61.73 & 71.74 & 57.52 & 48.46
 & 40.90 & 53.85 & 35.81 & 28.81
 & 37.46 & \underline{49.33} & 33.27 & 25.89 \\

\cellcolor{gray!10}
 CLIPLoRA+Fed-Grab~\cite{Fed-Grab}
 & \cellcolor{gray!10}\underline{70.24} & \cellcolor{gray!10}79.43 & \cellcolor{gray!10}66.75 & \cellcolor{gray!10}57.09
 & \cellcolor{gray!10}\underline{68.94} & \cellcolor{gray!10}\underline{77.14} & \cellcolor{gray!10}\underline{65.56} & \cellcolor{gray!10}\underline{55.56}
 & \cellcolor{gray!10}\underline{42.00} & \cellcolor{gray!10}53.56 & \cellcolor{gray!10}\underline{37.69} & \cellcolor{gray!10}\underline{30.92}
 & \cellcolor{gray!10}\underline{38.31} & \cellcolor{gray!10}49.17 & \cellcolor{gray!10}\underline{34.77} & \cellcolor{gray!10}\underline{27.16} \\

\cellcolor[HTML]{D7F6FF}\textbf{FedPuReL (ours)}
 & \cellcolor[HTML]{D7F6FF}\textbf{71.43} & \cellcolor[HTML]{D7F6FF}78.89 & \cellcolor[HTML]{D7F6FF}\textbf{67.69} & \cellcolor[HTML]{D7F6FF}\textbf{66.31}
 & \cellcolor[HTML]{D7F6FF}\textbf{70.55} & \cellcolor[HTML]{D7F6FF}76.09 & \cellcolor[HTML]{D7F6FF}\textbf{67.63} & \cellcolor[HTML]{D7F6FF}\textbf{66.98}
 & \cellcolor[HTML]{D7F6FF}\textbf{44.41} & \cellcolor[HTML]{D7F6FF}\textbf{60.72} & \cellcolor[HTML]{D7F6FF}\textbf{43.08} & \cellcolor[HTML]{D7F6FF}\textbf{35.79}
 & \cellcolor[HTML]{D7F6FF}\textbf{41.13} & \cellcolor[HTML]{D7F6FF}\textbf{52.58} & \cellcolor[HTML]{D7F6FF}\textbf{35.42} & \cellcolor[HTML]{D7F6FF}\textbf{32.83} \\

\textcolor{gray!60}&
\blueup{1.19} & \reddown{2.02} & \blueup{0.85} & \blueup{7.48} &
\blueup{4.48} & \reddown{1.73} & \blueup{2.07} & \blueup{11.42} &
\blueup{2.41} & \blueup{6.21} & \blueup{5.39} & \blueup{4.87} &
\blueup{2.82} & \blueup{3.25} & \blueup{0.65} & \blueup{5.67} \\

\hline
\multicolumn{17}{l}{\textcolor{gray!60}{\textit{Adapter-based}}} \\
\cellcolor{gray!10}
FedClip~\cite{lu2023fedclip}
 & \cellcolor{gray!10}68.97 & \cellcolor{gray!10}\underline{74.59} & \cellcolor{gray!10}65.67 & \cellcolor{gray!10}59.47
 & \cellcolor{gray!10}\underline{68.63} & \cellcolor{gray!10}\textbf{75.19} & \cellcolor{gray!10}\underline{65.43} & \cellcolor{gray!10}59.02
 & \cellcolor{gray!10}39.28 & \cellcolor{gray!10}\underline{47.40} & \cellcolor{gray!10}34.75 & \cellcolor{gray!10}34.66
 & \cellcolor{gray!10}38.59 & \cellcolor{gray!10}\textbf{47.13} & \cellcolor{gray!10}33.78 & \cellcolor{gray!10}33.74 \\

FedClip+Fed-Grab~\cite{Fed-Grab}
 & \underline{69.25} & \textbf{75.68} & \underline{66.23} & \underline{61.85}
 & 67.50 & 72.33 & 65.07 & \underline{62.81}
 & \underline{39.43} & 46.52 & \underline{35.34} & \underline{35.68}
 & \underline{38.95} & \underline{45.95} & \underline{34.81} & \underline{35.56} \\

\cellcolor[HTML]{D7F6FF}\textbf{FedPuReL (ours)}
 & \cellcolor[HTML]{D7F6FF}\textbf{70.51} & \cellcolor[HTML]{D7F6FF}73.52 & \cellcolor[HTML]{D7F6FF}\textbf{69.79} & \cellcolor[HTML]{D7F6FF}\textbf{67.49}
 & \cellcolor[HTML]{D7F6FF}\textbf{70.19} & \cellcolor[HTML]{D7F6FF}\underline{74.24} & \cellcolor[HTML]{D7F6FF}\textbf{69.27} & \cellcolor[HTML]{D7F6FF}\textbf{67.52}
 & \cellcolor[HTML]{D7F6FF}\textbf{41.99} & \cellcolor[HTML]{D7F6FF}\textbf{51.51} & \cellcolor[HTML]{D7F6FF}\textbf{39.00} & \cellcolor[HTML]{D7F6FF}\textbf{38.89}
 & \cellcolor[HTML]{D7F6FF}\textbf{40.12} & \cellcolor[HTML]{D7F6FF}45.48 & \cellcolor[HTML]{D7F6FF}\textbf{37.06} & \cellcolor[HTML]{D7F6FF}\textbf{36.39} \\

\textcolor{gray!60} &
\blueup{1.26} & \reddown{2.16} & \blueup{3.56} & \blueup{5.64} &
\blueup{1.56} & \reddown{0.95} & \blueup{3.84} & \blueup{4.71} &
\blueup{2.56} & \blueup{4.11} & \blueup{3.66} & \blueup{3.21} &
\blueup{1.17} & \reddown{1.65} & \blueup{2.25} & \blueup{0.83} \\

\end{tabular}
 \vspace{-1em}
\label{tab:imagenet_places_lt}
\end{table*}

\section{Experiments}
\subsection{Experimental Setup}
\noindent \textbf{Datasets.} We conduct experiments on Food101-LT~\cite{food101}, DTD-LT~\cite{dtd}, FGVC-Aricraft-LT~\cite{fgvc}, Stanford Dogs-LT~\cite{dogs}, OxfordPets-LT~\cite{pets}, and CIFAR-100-LT~\cite{cifar}, ImageNet-LT~\cite{deng2009imagenet}, Places-LT~\cite{zhou2017places} datasets. For the first six datasets, we follow \cite{cao2019learning} to create long-tailed versions using exponential decay with imbalance factor. For ImageNet-LT and Places-LT, we use the standard version from \cite{imagenet-lt}.

\noindent \textbf{Evaluation Protocol.} To evaluate global model (GM) performance, we calculate test accuracy using a globally balanced dataset and report accuracy across many ($>$100 samples), medium (20-100 samples), and few ($<$100 samples) classes. For personalized model (PM) evaluation, we employ local test accuracy, where each local test set is sampled from the global test set to maintain an identical distribution to its corresponding local training set. The PM accuracy is computed as the arithmetic mean of local test accuracies across all clients. Detailed categorization criteria for these class groups are provided in the Appendix.

\noindent \textbf{Baselines.} We compare FedPuReL against three representative PEFT categories: {\ding{182}} \textit{Prompt-based} , {\ding{183}} \textit{LoRA-based}, and {\ding{184}} \textit{Adapter-based}, including: PromptFL~\cite{guo2023promptfl}, PromptFolio~\cite{pan2024promptfolio}, FedOTP~\cite{li2024fedotp}, FedPGP~\cite{cui2024fedpgp}, CLIPLoRA~\cite{cliplora}, FedSA-LoRA~\cite{wang2024fedsalora} and FedClip~\cite{lu2023fedclip}.

To demonstrate FedPuReL's advantages over methods that explicitly address class imbalance, we additionally include Fed-GraB~\cite{Fed-Grab}, which uses gradient norm-based reweighting to mitigate head-class bias, combined with strong baselines (PromptFolio+Fed-GraB, CLIPLoRA+Fed-GraB, FedClip+Fed-GraB).

\noindent \textbf{Implementation Details.}
For a fair comparison, we follow the experimental settings used in prior work~\cite{guo2023promptfl,li2024fedotp,pan2024promptfolio,lu2023fedclip}. We employ CLIP with ViT-B/16 as the backbone. For prompt-based methods, the prompt length is set to 4, while for LoRA-based methods, the LoRA rank is set to 8. In our federated learning setup, we deploy 20 clients with non-IID data distribution controlled by Dirichlet parameter $\alpha=1$ and $\text{IF}=100$ by default. Training is conducted over 100 communication rounds using SGD optimizer, with 40\% of clients randomly selected to participate in each round.

\subsection{Comparison with State-of-the-Art}
\label{Comparison with State-of-the-Art}

\begin{table}[!t]\footnotesize
\caption{\textbf{Comparison with state-of-the-art methods} on CIFAR-100-LT under varying imbalance factors for global models (GM) and personalized models (PM).}
\centering
\scriptsize{
\resizebox{\linewidth}{!}{
\setlength\tabcolsep{1.5pt}
\renewcommand\arraystretch{1.1}
\begin{tabular}{r||cc|cc|cc|cc}
\cline{2-9}
\hline\thickhline
\rowcolor{gray!20}
\rowcolor{gray!20}
\textbf{CIFAR-100-LT} &
  \multicolumn{2}{c|}{\textbf{IF=500}} &
  \multicolumn{2}{c|}{\textbf{IF=200}} &
  \multicolumn{2}{c|}{\textbf{IF=100}} &
  \multicolumn{2}{c}{\textbf{IF=50}} \\
\cline{2-9}
\rowcolor{gray!20}
\textbf{Method} & \textbf{GM} & \textbf{PM} & \textbf{GM} & \textbf{PM} & \textbf{GM} & \textbf{PM} & \textbf{GM} & \textbf{PM} \\
\hline\hline
Zero-shot &64.82 & 64.79 & 64.82 & 64.79 & 64.82 & 64.79 & 64.82 & 64.79 \\
\hdashline
\multicolumn{9}{l}{\textcolor{gray!60}{\textit{Prompt-based}}} \\
\cellcolor{gray!10}PromptFL~\cite{guo2023promptfl}
 & \cellcolor{gray!10}56.12 & \cellcolor{gray!10}\underline{67.50} & \cellcolor{gray!10}61.21 & \cellcolor{gray!10}\underline{71.90}
 & \cellcolor{gray!10}61.98 & \cellcolor{gray!10}70.52 & \cellcolor{gray!10}64.50 & \cellcolor{gray!10}72.51 \\

PromptFolio~\cite{pan2024promptfolio}
 & 57.36 & 63.80 & 60.58 & 69.74
 & 64.10 & 70.54 & 67.15 & 72.69 \\
\cellcolor{gray!10}FedOTP~\cite{li2024fedotp}
 & \cellcolor{gray!10}52.59 & \cellcolor{gray!10}58.43 & \cellcolor{gray!10}51.33 & \cellcolor{gray!10}65.09
 & \cellcolor{gray!10}58.02 & \cellcolor{gray!10}65.23 & \cellcolor{gray!10}61.30 & \cellcolor{gray!10}68.64 \\

FedPGP~\cite{cui2024fedpgp}
 & 55.88 & 62.69 & 59.48 & 67.47
 & 62.96 & 67.83 & 65.55 & 71.23 \\
 
 \cellcolor{gray!10}PromptFolio+Fed-GraB~\cite{Fed-Grab}
 & \cellcolor{gray!10}\underline{58.98} & \cellcolor{gray!10}65.28 & \cellcolor{gray!10}\underline{62.18} & \cellcolor{gray!10}71.00
 & \cellcolor{gray!10}\underline{65.14} & \cellcolor{gray!10}\underline{71.79} & \cellcolor{gray!10}\underline{68.49} & \cellcolor{gray!10}\underline{73.60} \\

\cellcolor[HTML]{D7F6FF}\textbf{FedPuReL (ours)}
 & \cellcolor[HTML]{D7F6FF}\textbf{67.20} & \cellcolor[HTML]{D7F6FF}\textbf{72.70} & \cellcolor[HTML]{D7F6FF}\textbf{68.01} & \cellcolor[HTML]{D7F6FF}\textbf{74.67}
 & \cellcolor[HTML]{D7F6FF}\textbf{69.77} & \cellcolor[HTML]{D7F6FF}\textbf{73.37} & \cellcolor[HTML]{D7F6FF}\textbf{70.40} & \cellcolor[HTML]{D7F6FF}\textbf{74.68} \\
 
\textcolor{gray!60}&
\blueup{9.84} & \blueup{5.20} &
\blueup{5.83} & \blueup{2.77} &
\blueup{4.63} & \blueup{1.58} &
\blueup{1.91} & \blueup{1.08} \\
\hline
\multicolumn{9}{l}{\textcolor{gray!60}{\textit{LoRA-based}}} \\
\cellcolor{gray!10}CLIPLoRA~\cite{cliplora}
 & \cellcolor{gray!10}69.37 & \cellcolor{gray!10}73.68 & \cellcolor{gray!10}71.71 & \cellcolor{gray!10}76.63
 & \cellcolor{gray!10}75.02 & \cellcolor{gray!10}77.80 & \cellcolor{gray!10}76.85 & \cellcolor{gray!10}79.52 \\

FedSA-LoRA~\cite{wang2024fedsalora}
 & 66.51 & \underline{74.26} & 71.66 & 76.50
 & 74.48 & \underline{81.68} & 72.86 & 78.61 \\
 
\cellcolor{gray!10} CLIPLoRA+Fed-GraB~\cite{Fed-Grab}
 & \cellcolor{gray!10}\underline{70.32} & \cellcolor{gray!10}74.17 & \cellcolor{gray!10}\underline{72.58} & \cellcolor{gray!10}\underline{77.43}
 & \cellcolor{gray!10}\underline{76.35} & \cellcolor{gray!10}77.91 & \cellcolor{gray!10}\underline{77.47} & \cellcolor{gray!10}\underline{80.83} \\

\cellcolor[HTML]{D7F6FF}\textbf{FedPuReL (ours)}
 & \cellcolor[HTML]{D7F6FF}\textbf{72.91} & \cellcolor[HTML]{D7F6FF}\textbf{77.73} & \cellcolor[HTML]{D7F6FF}\textbf{74.26} & \cellcolor[HTML]{D7F6FF}\textbf{80.04}
 & \cellcolor[HTML]{D7F6FF}\textbf{77.56} & \cellcolor[HTML]{D7F6FF}\textbf{84.62} & \cellcolor[HTML]{D7F6FF}\textbf{78.12} & \cellcolor[HTML]{D7F6FF}\textbf{81.85} \\
\textcolor{gray!60} &
\blueup{2.59} & \blueup{3.47} &
\blueup{1.68} & \blueup{2.61} &
\blueup{1.21} & \blueup{2.94} &
\blueup{0.65} & \blueup{1.02} \\

\hline
\multicolumn{9}{l}{\textcolor{gray!60}{\textit{Adapter-based}}} \\
\cellcolor{gray!10}FedClip~\cite{lu2023fedclip}
& \cellcolor{gray!10}61.49 & \cellcolor{gray!10}65.77 & \cellcolor{gray!10}62.68 & \cellcolor{gray!10}68.61
 & \cellcolor{gray!10}64.09 & \cellcolor{gray!10}66.87 & \cellcolor{gray!10}65.59 & \cellcolor{gray!10}68.68 \\

 FedClip+Fed-GraB~\cite{Fed-Grab}
 & \underline{62.51} & \underline{67.14} & \underline{63.86} & \underline{69.48}
 & \underline{65.40} & \underline{67.20} & \underline{66.60} & \textbf{69.99} \\
 
\cellcolor[HTML]{D7F6FF}\textbf{FedPuReL (ours)}
 & \cellcolor[HTML]{D7F6FF}\textbf{66.52} & \cellcolor[HTML]{D7F6FF}\textbf{68.44} & \cellcolor[HTML]{D7F6FF}\textbf{66.55} & \cellcolor[HTML]{D7F6FF}\textbf{69.84}
 & \cellcolor[HTML]{D7F6FF}\textbf{67.54} & \cellcolor[HTML]{D7F6FF}\textbf{68.42} & \cellcolor[HTML]{D7F6FF}\textbf{67.13} & \cellcolor[HTML]{D7F6FF}\underline{69.89} \\
 
\textcolor{gray!60}&
\blueup{4.01} & \blueup{1.30} &
\blueup{2.69} & \blueup{0.36} &
\blueup{2.14} & \blueup{1.22} &
\blueup{0.53} & \reddown{0.10} \\

\end{tabular}}
}
\vspace{-12pt}
\label{tab:cifar100_lt}
\end{table}

\begin{table*}[t]
\centering
\caption{\textbf{Comparison with state-of-the-art methods} on Food101-LT, DTD-LT, Aircraft-LT, Dogs-LT, and Pets-LT datasets.}
\vspace{-0.5em}
\footnotesize
\setlength\tabcolsep{3pt}
\renewcommand\arraystretch{1.1}
\begin{tabular}{r||cc|cc|cc|cc|cc}
\hline\thickhline
\rowcolor{gray!20}
\multirow{2}{*}{\textbf{Method}} & \multicolumn{2}{c|}{\textbf{Food101-LT}} & \multicolumn{2}{c|}{\textbf{DTD-LT}} & \multicolumn{2}{c|}{\textbf{Aircraft-LT}} & \multicolumn{2}{c|}{\textbf{Dogs-LT}} & \multicolumn{2}{c}{\textbf{Pets-LT}} \\
\cline{2-11}
\rowcolor{gray!20}
\multirow{-2}{*}{\textbf{Method}}  & \textbf{GM} & \textbf{PM} & \textbf{GM} & \textbf{PM} & \textbf{GM} & \textbf{PM} & \textbf{GM} & \textbf{PM} & \textbf{GM} & \textbf{PM} \\
\hline\hline
Zero-shot & 85.39 & 85.53 & 48.41 & 41.46 & 21.54 & 29.38 & 66.43 & 60.19 & 90.41 & 86.03 \\
\hdashline
\multicolumn{11}{l}{\textcolor{gray!60}{\textit{Prompt-based}}} \\
\cellcolor{gray!10}PromptFL~\cite{guo2023promptfl}
 & \cellcolor{gray!10}82.74 & \cellcolor{gray!10}\underline{88.06} & \cellcolor{gray!10}46.22 & \cellcolor{gray!10}63.49
 & \cellcolor{gray!10}26.59 & \cellcolor{gray!10}28.51 & \cellcolor{gray!10}62.57 & \cellcolor{gray!10}67.61
 & \cellcolor{gray!10}86.80 & \cellcolor{gray!10}91.98 \\

PromptFolio~\cite{pan2024promptfolio}
 & 81.49 & 87.09 & 54.97 & 70.02
 & 29.26 & \underline{34.68} & \underline{64.94} & \underline{70.30}
 & 85.15 & 91.08 \\

\cellcolor{gray!10}FedOTP~\cite{li2024fedotp}
 & \cellcolor{gray!10}74.61 & \cellcolor{gray!10}84.52 & \cellcolor{gray!10}48.70 & \cellcolor{gray!10}68.63
 & \cellcolor{gray!10}22.50 & \cellcolor{gray!10}24.55 & \cellcolor{gray!10}64.76 & \cellcolor{gray!10}65.17
 & \cellcolor{gray!10}67.46 & \cellcolor{gray!10}83.73 \\

FedPGP~\cite{cui2024fedpgp}
 & 81.43 & 85.91 & 48.47 & 64.96
 & 24.81 & 26.70 & 62.44 & 65.85
 & 84.85 & 92.00 \\
\cellcolor{gray!10}
PromptFolio+Fed-GraB~\cite{Fed-Grab}
 & \cellcolor{gray!10}\underline{83.59} & \cellcolor{gray!10}88.03 & \cellcolor{gray!10}\underline{56.21} & \cellcolor{gray!10}\underline{70.21}
 & \cellcolor{gray!10}\underline{29.91} & \cellcolor{gray!10}\textbf{35.12} & \cellcolor{gray!10}63.52 & \cellcolor{gray!10}61.03
 & \cellcolor{gray!10}\underline{87.73} & \cellcolor{gray!10}\underline{91.96} \\

\cellcolor[HTML]{D7F6FF}\textbf{FedPuReL (ours)}
 & \cellcolor[HTML]{D7F6FF}\textbf{86.71}\blueupsmall{3.12} & \cellcolor[HTML]{D7F6FF}\textbf{89.26}\blueupsmall{1.20} & \cellcolor[HTML]{D7F6FF}\textbf{57.75}\blueupsmall{1.54} & \cellcolor[HTML]{D7F6FF}\textbf{70.46}\blueupsmall{0.25}
 & \cellcolor[HTML]{D7F6FF}\textbf{32.81}\blueupsmall{2.90} & \cellcolor[HTML]{D7F6FF}34.15\reddownsmall{0.97} & \cellcolor[HTML]{D7F6FF}\textbf{67.06}\blueupsmall{2.12} & \cellcolor[HTML]{D7F6FF}\textbf{72.48}\blueupsmall{2.18}
 & \cellcolor[HTML]{D7F6FF}\textbf{92.13}\blueupsmall{4.40} & \cellcolor[HTML]{D7F6FF}\textbf{94.53}\blueupsmall{2.57} \\
 
\hline
\multicolumn{11}{l}{\textcolor{gray!60}{\textit{LoRA-based}}} \\
\cellcolor{gray!10}
CLIPLoRA~\cite{cliplora}
 & \cellcolor{gray!10}82.27 & \cellcolor{gray!10}87.43 & \cellcolor{gray!10}51.76 & \cellcolor{gray!10}66.83
 & \cellcolor{gray!10}30.02 & \cellcolor{gray!10}\underline{33.61} & \cellcolor{gray!10}67.49 & \cellcolor{gray!10}\underline{71.28}
 & \cellcolor{gray!10}88.77 & \cellcolor{gray!10}93.56 \\

FedSA-LoRA~\cite{wang2024fedsalora}
 & 83.40 & 86.50 & 44.53 & 68.60
 & 26.31 & 31.23 & 66.10 & 67.08
 & 88.05 & 93.71 \\

\cellcolor{gray!10}
CLIPLoRA+Fed-GraB~\cite{Fed-Grab}
 & \cellcolor{gray!10}\underline{84.84} & \cellcolor{gray!10}\underline{87.69} & \cellcolor{gray!10}\underline{54.00} & \cellcolor{gray!10}\underline{69.07}
 & \cellcolor{gray!10}\underline{31.62} & \cellcolor{gray!10}33.35 & \cellcolor{gray!10}\underline{68.29} & \cellcolor{gray!10}69.63
 & \cellcolor{gray!10}\underline{89.29} & \cellcolor{gray!10}\underline{93.91} \\
 
\cellcolor[HTML]{D7F6FF}\textbf{FedPuReL (ours)}
 & \cellcolor[HTML]{D7F6FF}\textbf{86.63}\blueupsmall{1.78} & \cellcolor[HTML]{D7F6FF}\textbf{89.32}\blueupsmall{1.63} & \cellcolor[HTML]{D7F6FF}\textbf{54.45}\blueupsmall{0.45} & \cellcolor[HTML]{D7F6FF}\textbf{69.56}\blueupsmall{0.49}
 & \cellcolor[HTML]{D7F6FF}\textbf{32.68}\blueupsmall{1.06} & \cellcolor[HTML]{D7F6FF}\textbf{37.02}\blueupsmall{3.41} & \cellcolor[HTML]{D7F6FF}\textbf{69.25}\blueupsmall{0.96} & \cellcolor[HTML]{D7F6FF}\textbf{73.54}\blueupsmall{2.26}
 & \cellcolor[HTML]{D7F6FF}\textbf{91.98}\blueupsmall{2.69} & \cellcolor[HTML]{D7F6FF}\textbf{94.60}\blueupsmall{0.69} \\

\hline
\multicolumn{11}{l}{\textcolor{gray!60}{\textit{Adapter-based}}} \\
\cellcolor{gray!10}
FedClip~\cite{lu2023fedclip}
& \cellcolor{gray!10}83.75 
& \cellcolor{gray!10}85.76 
& \cellcolor{gray!10}\underline{46.01} 
& \cellcolor{gray!10}\underline{50.47} 
& \cellcolor{gray!10}\underline{22.83} 
& \cellcolor{gray!10}22.66 
& \cellcolor{gray!10}\underline{63.74} 
& \cellcolor{gray!10}\underline{61.10} 
& \cellcolor{gray!10}89.24 
& \cellcolor{gray!10}90.55
\\
 
FedClip+Fed-GraB~\cite{Fed-Grab}
& \underline{84.35} & \underline{86.14} & 44.60 & 49.33
& 22.59 & \underline{22.78} & 63.52 & 61.03
& \underline{90.01} & \underline{91.63} \\
 
\cellcolor[HTML]{D7F6FF}\textbf{FedPuReL (ours)}
& \cellcolor[HTML]{D7F6FF}\textbf{85.76}\blueupsmall{1.41}
& \cellcolor[HTML]{D7F6FF}\textbf{87.08}\blueupsmall{0.94}
& \cellcolor[HTML]{D7F6FF}\textbf{48.15}\blueupsmall{2.14}
& \cellcolor[HTML]{D7F6FF}\textbf{51.52}\blueupsmall{1.05}
& \cellcolor[HTML]{D7F6FF}\textbf{24.78}\blueupsmall{1.95}
& \cellcolor[HTML]{D7F6FF}\textbf{24.16}\blueupsmall{1.38}
& \cellcolor[HTML]{D7F6FF}\textbf{66.01}\blueupsmall{2.27}
& \cellcolor[HTML]{D7F6FF}\textbf{62.10}\blueupsmall{1.00}
& \cellcolor[HTML]{D7F6FF}\textbf{90.98}\blueupsmall{0.97}
& \cellcolor[HTML]{D7F6FF}\textbf{92.76}\blueupsmall{1.13} \\

\end{tabular}
\label{tab:other_datasets_lt}
\end{table*}

\begin{table*}[!t]
\centering
\caption{\textbf{Comparison on CIFAR100 and Places-LT} under Global (G) and Personalized (P) settings. Best in \textbf{bold}, second-best \underline{underlined}.}
\vspace{-0.5em}
\footnotesize
\setlength\tabcolsep{2.5pt}
\renewcommand\arraystretch{1.1}
\begin{tabular}{l||cc|cccc|cccc|cccc|cccc}
\hline\thickhline
\rowcolor{gray!20}
\multirow{3}{*}{\textbf{Method}} & \multirow{3}{*}{\textbf{G}}& \multirow{3}{*}{\textbf{P}} & \multicolumn{8}{c|}{\textbf{CIFAR-100-LT}} & \multicolumn{8}{c}{\textbf{Places-LT}} \\
\cline{4-19}
\rowcolor{gray!20}
& & & \multicolumn{4}{c}{\textbf{GM}} & \multicolumn{4}{c|}{\textbf{PM}} & \multicolumn{4}{c}{\textbf{GM}} & \multicolumn{4}{c}{\textbf{PM}} \\
\cline{4-19}
\rowcolor{gray!20}
\multirow{-3}{*}{\textbf{Method}} & \multirow{-3}{*}{\textbf{G}}& \multirow{-3}{*}{\textbf{P}} & \textbf{All} & \textbf{Many} & \textbf{Med.} & \textbf{Few}
& \textbf{All} & \textbf{Many} & \textbf{Med.} & \textbf{Few}
& \textbf{All} & \textbf{Many} & \textbf{Med.} & \textbf{Few}
& \textbf{All} & \textbf{Many} & \textbf{Med.} & \textbf{Few} \\
\hline
Baseline & \xmark & \xmark & 61.98  & 80.23 & 61.12 & 42.32 & 70.52 & 82.53 & 62.15 & 48.67 & 38.90 & 51.07 & 32.86 & 29.64 & 36.49 & 48.00 & 31.37 & 27.42 \\
+ Gradient Purification~~~~~ & \cmark & \textcolor{\tblgray}{\xmark} & 69.77 & 78.63 & 67.08 & 64.74 & 72.06 & 81.91 & 63.64 & 57.46 & 43.88 & 55.44 & 40.94 & 39.06 & 38.14 & 48.34 & 35.27 & 32.72 \\
\grayrow\textcolor{\tblgray}{~~~\textit{v1}. LS-KD~{\hypersetup{citecolor=\tblgray}\cite{sun2024logit}}} & {\cmark} & \textcolor{\tblgray}{\xmark} & \textcolor{\tblgray}{66.43} & \textcolor{\tblgray}{80.77} & \textcolor{\tblgray}{64.29} & \textcolor{\tblgray}{52.58} & \textcolor{\tblgray}{71.56} & \textcolor{\tblgray}{81.59} & \textcolor{\tblgray}{63.32} & \textcolor{\tblgray}{55.26} & \textcolor{\tblgray}{40.42} & \textcolor{\tblgray}{50.58} & \textcolor{\tblgray}{34.34} & \textcolor{\tblgray}{35.49} & \textcolor{\tblgray}{37.47} & \textcolor{\tblgray}{48.22} & \textcolor{\tblgray}{33.95} & \textcolor{\tblgray}{32.89} \\
    \grayrow\textcolor{\tblgray}{~~~\textit{v2}. WiSE-FT~{\hypersetup{citecolor=\tblgray}\cite{wortsman2022robust}}} & {\cmark} & \textcolor{\tblgray}{\xmark} & \textcolor{\tblgray}{63.64} & \textcolor{\tblgray}{78.86} & \textcolor{\tblgray}{62.03} & \textcolor{\tblgray}{49.42} & \textcolor{\tblgray}{69.94} & \textcolor{\tblgray}{80.92} & \textcolor{\tblgray}{62.48} & \textcolor{\tblgray}{50.11} & \textcolor{\tblgray}{38.95} & \textcolor{\tblgray}{48.67} & \textcolor{\tblgray}{33.72} & \textcolor{\tblgray}{39.14} & \textcolor{\tblgray}{37.04} & \textcolor{\tblgray}{44.81} & \textcolor{\tblgray}{32.52} & \textcolor{\tblgray}{32.90} \\
\hline
+ Residual & \textcolor{\tblgray}{\xmark} & \cmark & 68.43 & 78.80 & 65.82 & 59.58 & 72.25 & 82.28 & 63.30 & 55.89 & 42.27 & 49.84 & 37.25 & 39.64 & 39.34 & 48.75 & 33.88 & 32.35 \\
\grayrow\textcolor{\tblgray}{~~~\textit{v1}. PromptFolio~{\hypersetup{citecolor=\tblgray}\cite{pan2024promptfolio}}} & \textcolor{\tblgray}{\xmark} & {\cmark} & \textcolor{\tblgray}{65.90} & \textcolor{\tblgray}{83.74} & \textcolor{\tblgray}{66.82} & \textcolor{\tblgray}{44.74} & \textcolor{\tblgray}{69.91} & \textcolor{\tblgray}{84.93} & \textcolor{\tblgray}{62.66} & \textcolor{\tblgray}{46.49} & \textcolor{\tblgray}{41.30} & \textcolor{\tblgray}{52.90} & \textcolor{\tblgray}{36.47} & \textcolor{\tblgray}{31.01} & \textcolor{\tblgray}{37.14} & \textcolor{\tblgray}{47.69} & \textcolor{\tblgray}{34.20} & \textcolor{\tblgray}{28.58} \\
\grayrow\textcolor{\tblgray}{~~~\textit{v2}. FedOTP~{\hypersetup{citecolor=\tblgray}\cite{li2024fedotp}}} & \textcolor{\tblgray}{\xmark} & {\cmark} & \textcolor{\tblgray}{62.96} & \textcolor{\tblgray}{84.03} & \textcolor{\tblgray}{63.72} & \textcolor{\tblgray}{41.68} & \textcolor{\tblgray}{66.53} & \textcolor{\tblgray}{83.36} & \textcolor{\tblgray}{58.3} & \textcolor{\tblgray}{35.49} & \textcolor{\tblgray}{36.37} & \textcolor{\tblgray}{48.73} & \textcolor{\tblgray}{32.27} & \textcolor{\tblgray}{23.07} & \textcolor{\tblgray}{34.25} & \textcolor{\tblgray}{46.03} & \textcolor{\tblgray}{30.41} & \textcolor{\tblgray}{22.12} \\
\rowcolor[HTML]{D7F6FF}
\textbf{Ours} & \cmark & \cmark & \textbf{69.77} & \textbf{78.63} & \textbf{67.08} & \textbf{64.74} & \textbf{73.37} & \textbf{82.67} & \textbf{65.69} & \textbf{58.25} & \textbf{43.88} & \textbf{55.44} & \textbf{40.94} & \textbf{39.06} & \textbf{39.68} & \textbf{49.85} & \textbf{34.55} & \textbf{35.56} \\
\end{tabular}
\parbox{\linewidth}{\footnotesize \vspace{0.3em} \textbf{\textit{v1}/\textit{v2}}: Contender for initialization and regularization components.}
\label{ablation}
\vspace{-12pt}
\end{table*}

\noindent \textbf{Performance on Large-Scale Benchmarks.}
Table~\ref{tab:imagenet_places_lt} compares FedPuReL on ImageNet-LT and Places-LT. FedPuReL consistently outperforms all baselines in both GM and PM settings, particularly on few classes, even exceeding Fed-GraB-enhanced variants. This performance pattern aligns with our design principle: gradient purification maintains zero-shot balance by preventing the global model from overfitting to head classes, instead reallocating optimization capacity toward under-represented categories. As evidenced by Fig.~\ref{fig:exp2} and the analysis in Sec.~\ref{Analysis of FedPuReL}, this mechanism yields substantially improved tail performance and higher balancedness compared to baseline. Table~\ref{tab:other_datasets_lt} demonstrates robustness across five additional datasets spanning fine-grained recognition, textures, and scenes.

\noindent \textbf{Performance Under Varying Imbalance Factors.}
Table~\ref{tab:cifar100_lt} evaluates FedPuReL on CIFAR-100-LT across varying imbalance factors. FedPuReL maintains superior performance as imbalance intensifies, with advantages growing proportionally, demonstrating effectiveness in severe long-tail scenarios. Fed-GraB's limited gains validate that relying solely on prior estimation is insufficient under severe long-tail scenarios, whereas our gradient purification naturally maintains balance by anchoring to zero-shot predictions.

\subsection{Ablation Study}
\label{Ablation Studies}

To thoroughly assess our approach, we conduct ablation studies investigating the efficacy of each component under long-tailed scenarios. We perform ablation experiments on CIFAR-100-LT and Places-LT datasets.

\noindent \textbf{Balancing Strategies for Global Training.}
Starting from a baseline without balancing mechanisms, we compare our method against alternative zero-shot guidance approaches:
\begin{itemize}
    \item \textit{LS-KD}~\cite{sun2024logit}: Apply Z-score standardization to logits before computing softmax and KL divergence.
    \item \textit{WiSE-FT}~\cite{wortsman2022robust}: Ensemble zero-shot and fine-tuned models in weight space by linearly interpolating with mixing coefficient $\alpha \in [0, 1]$ to improve OOD robustness.
    
\end{itemize}
As shown in Table~\ref{ablation}, our \textit{Gradient Purification} achieves superior performance, demonstrating that projecting task gradients orthogonal to alignment gradients effectively preserves CLIP's balanced knowledge during task adaptation.

\noindent \textbf{Personalization Strategies.}
We evaluate our residual learning against existing personalized fusion approaches:
\begin{itemize}
    \item \textit{PromptFolio}~\cite{pan2024promptfolio}: Employ portfolio-based prompt selection via feature-level interpolation.
    \item \textit{FedOTP}~\cite{li2024fedotp}: Employ unbalanced optimal transport to align local visual features with global and local prompts.
\end{itemize}
Our \textit{Residual Learning}, even without gradient purification, surpasses both alternatives. This demonstrates that output-space residuals effectively mitigate bias inheritance.

Integrating gradient purification with residual learning yields gains exceeding those of individual components, demonstrating their synergy. The combination achieves the best results across both datasets. These findings confirm our thesis that jointly addressing balancedness and personalization is essential for effective federated long-tailed learning.

\subsection{More Supporting Experiments}
Due to the page limitation, we report additional results and ablation studies in Appendix~\ref{app:Additional Experiments}, including 1) different client heterogeneities; 2) ablation hyper-parameter $\lambda$ in Eq.~\ref{fusion loss}; 3) analysis of gradient angles; 4) model convergence.

\subsection{Analysis of FedPuReL}
\label{Analysis of FedPuReL}

\begin{figure}[!t]
\centering
\begin{subfigure}[b]{0.49\linewidth}
    \centering
    \includegraphics[width=\linewidth]{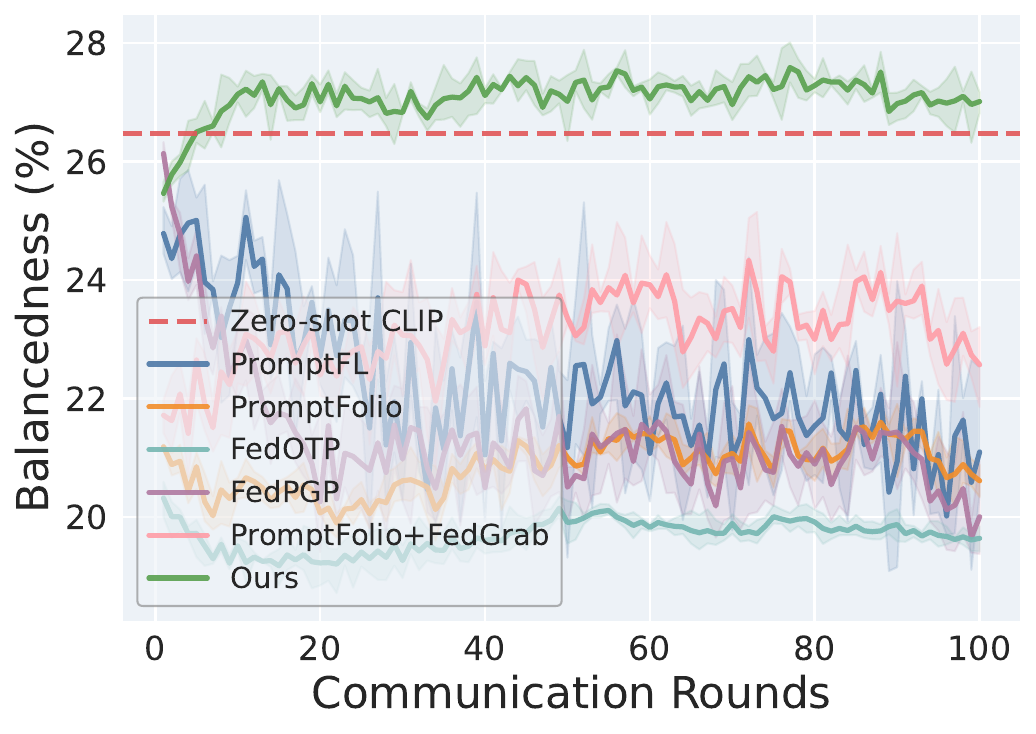}
    
    \caption*{(a) Global Model}
\end{subfigure}
\hfill
\begin{subfigure}[b]{0.49\linewidth}
    \centering
    \includegraphics[width=\linewidth]{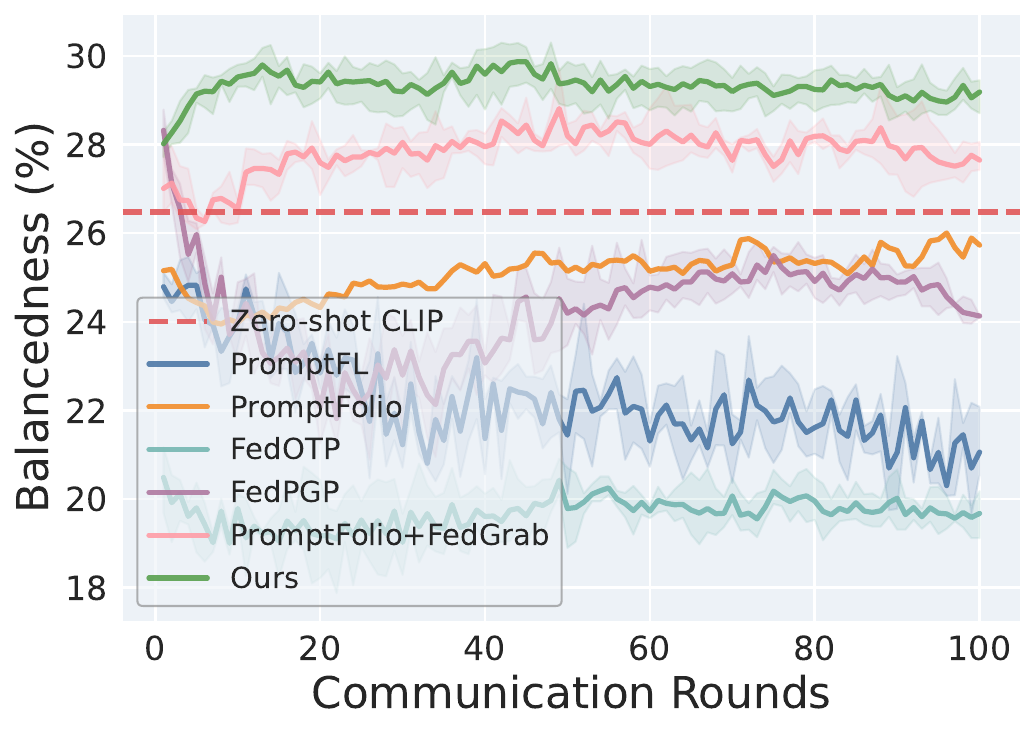}
    \caption*{(b) Personalized models}
\end{subfigure}
\vspace{-0.5em}
\caption{\textbf{Comparison of Balancedness} of SOTA prompt-based methods on global and personalized models for CIFAR-100-LT.}
\label{fig:exp2}
\end{figure}

\noindent \textbf{\hypertarget{analysis:1}{Analysis 1}: FedPuReL effectively maintains class balance.}
To assess whether FedPuReL preserves balance, we track the balancedness metric $\beta(V)$ throughout training (Fig.~\ref{fig:exp2}). Global models trained with standard FL methods exhibit significant balance degradation over time. In contrast, FedPuReL maintains close to the initial zero-shot level, demonstrating effective balance preservation through gradient purification. For personalized models, standard approaches inherit and propagate global bias, resulting in poor balancedness. FedPuReL substantially improves balancedness, validating that residual learning enables client-specific personalization without bias propagation. The minimal gap between global and personalized balancedness in FedPuReL, demonstrates that our approach uniformly benefits all clients across diverse data distributions, rather than trading global fairness for personalization gains.
\noindent \textbf{\hypertarget{analysis:2}{Analysis 2}: TKL metric quantifies divergence from zero-shot knowledge.} We employ our TKL metric, which neutralizes confidence variations to isolate true distributional changes  (Fig.~\ref{fig:motivation3}). Standard KL divergence exhibits large values even on balanced data due to confidence shifts, obscuring genuine bias. In contrast, TKL distinguishes balanced adaptation from biased divergence. Comparing results across datasets reveals that TKL divergence is higher on long-tailed data, quantifying how class imbalance systematically drives models away from CLIP's balanced knowledge. The strong negative correlation between TKL and $\beta(V)$ (Fig.~\ref{fig:motivation2}) confirms this relationship.

\begin{figure}[!t]
\centering
\begin{subfigure}[b]{0.49\linewidth}
  \centering
  \includegraphics[width=\linewidth]{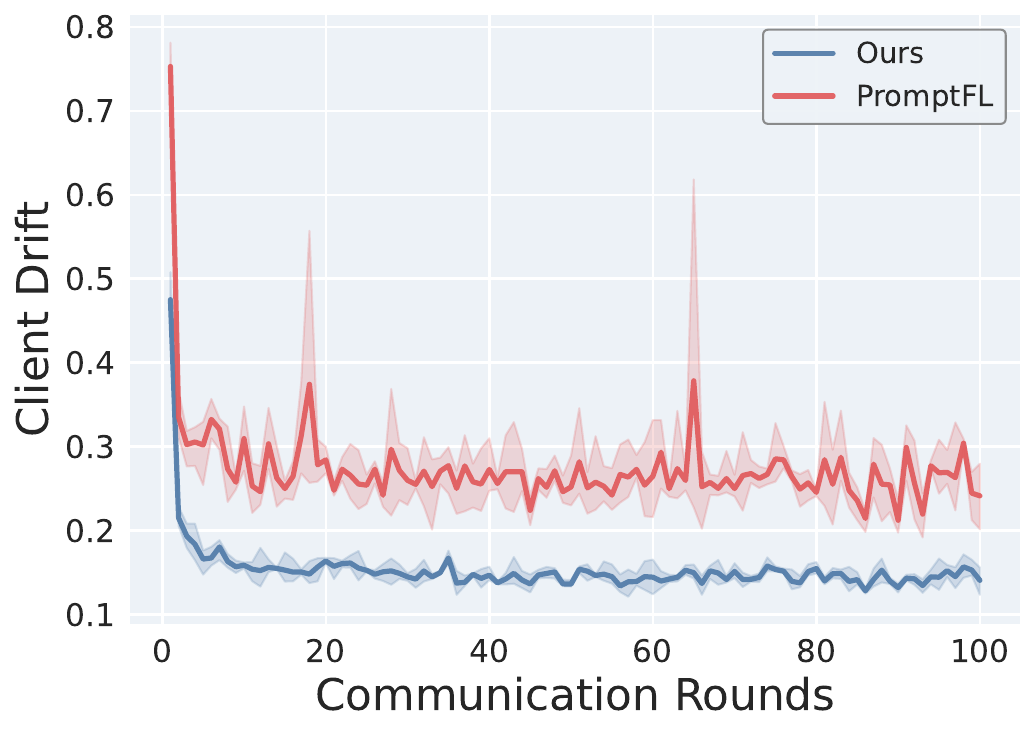}
  \subcaption{Client drift evolution}\label{fig:any1}
\end{subfigure}\hfill
\begin{subfigure}[b]{0.49\linewidth}
  \centering
  \includegraphics[width=\linewidth]{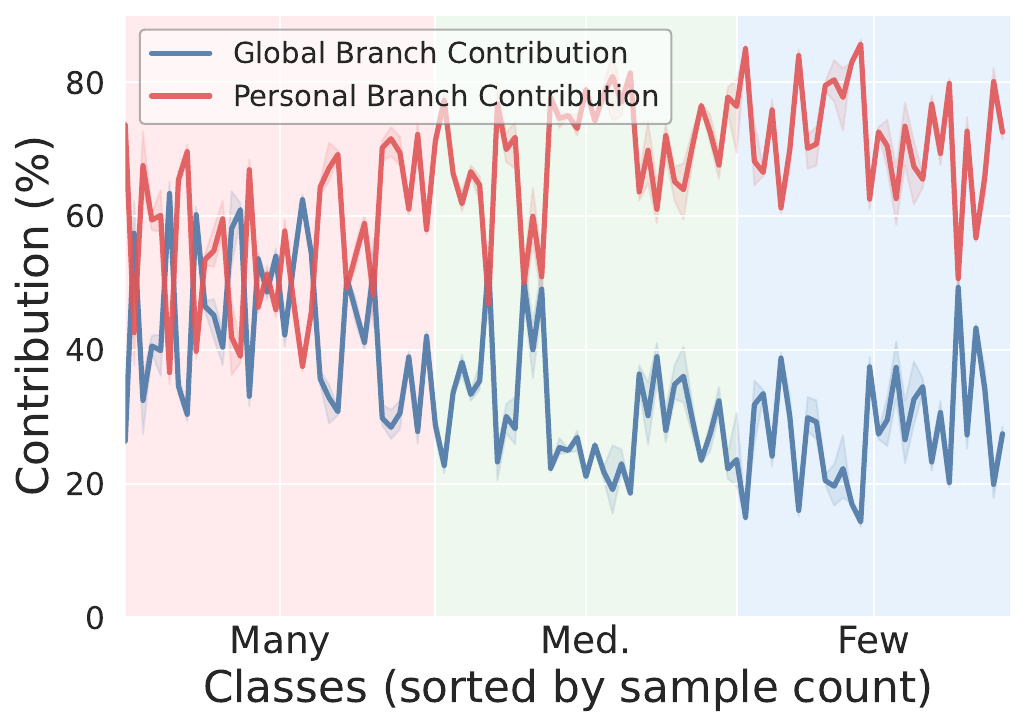}
  \subcaption{Branch contribution}\label{fig:any2}
\end{subfigure}
\caption{\textbf{Client drift and branch contribution analysis.} \textbf{(a)} Client drift measured as the L2 distance between local and global model. \textbf{(b)} Percentage contribution of global versus personalized branches to correct predictions, decomposed across classes sorted. Both analyses are conducted on CIFAR-100-LT.}
\label{fig:any}
\vspace{-12pt}
\end{figure}

\begin{figure}[!t]
\centering
\begin{subfigure}[b]{0.49\linewidth}
    \centering
    \includegraphics[width=\linewidth]{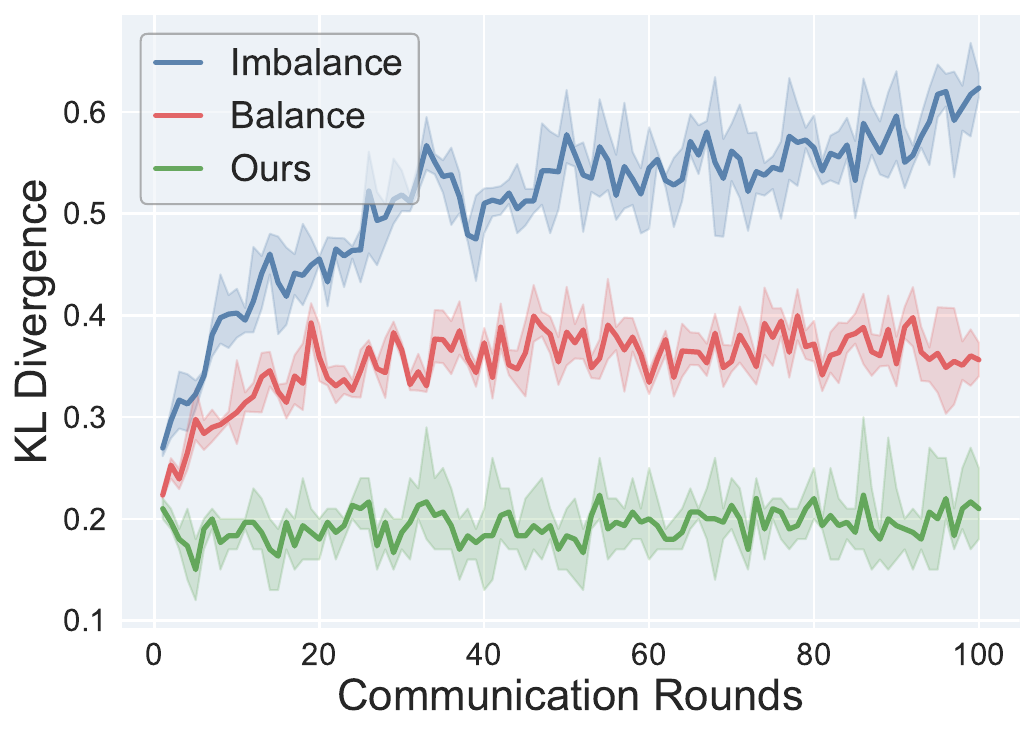}
    \caption*{(a) Temperature-aligned KL}
\end{subfigure}
\hfill
\begin{subfigure}[b]{0.49\linewidth}
    \centering
    \includegraphics[width=\linewidth]{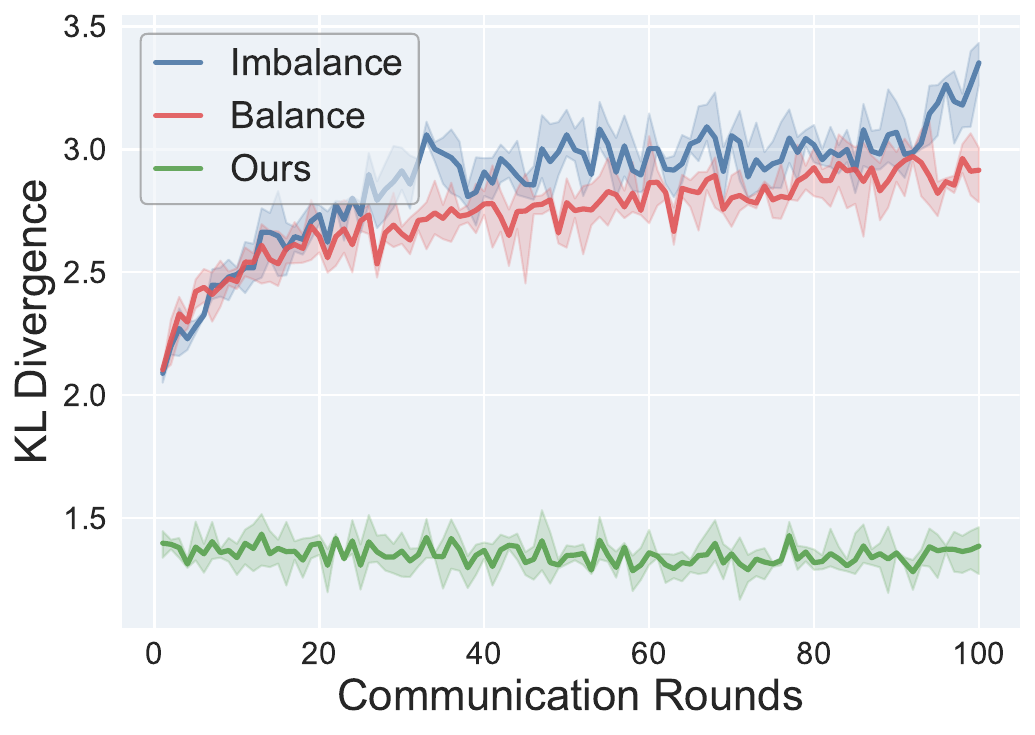}
    \caption*{(b) KL}
\end{subfigure}
\vspace{-0.5em}
\caption{\textbf{Comparison of different KL divergence metrics} across imbalanced and balanced CIFAR-100 datasets.}
\label{fig:motivation3}
\vspace{-12pt}
\end{figure}

\noindent \textbf{\hypertarget{analysis:3}{Analysis 3}: FedPuReL effectively reduces client drift.} 
Intuitively, when different clients purify their gradients toward the same zero-shot anchor, this shared reference point should improve cross-client consistency. To evaluate this, we measure client drift as the L2 distance between local and global parameters: $\text{drift}_k^{(t)} = \|\theta_k^{(t)} - \theta_{\text{global}}^{(t)}\|_2$. As illustrated in Fig.~\ref{fig:any1}, FedPuReL maintains substantially lower drift than PromptFL with minimal variance. In contrast, PromptFL shows volatile spikes, suggesting episodic severe divergence where some clients' updates conflict sharply with the global model. These results demonstrate that gradient purification maintains cross-client consistency with the global model, enabling faster convergence through improved aggregation quality.

\noindent \textbf{\hypertarget{analysis:4}{Analysis 4}: Global and personalized branches exhibit complementary contributions.} 
Our design separates balanced global knowledge from personalized local adaptation, which should lead to different contribution patterns across class types. Specifically, head classes with abundant samples should rely more on global knowledge, while tail classes with scarce data should require stronger personalized corrections. To evaluate this, we decompose each correct prediction to identify the dominant contributor between the global and personalized branches. As illustrated in Fig.~\ref{fig:any2}, as class frequency decreases from head to tail, the personalized branch contribution increases, while the global branch contribution decreases correspondingly. Notably, even for tail classes where personalization dominates, the global branch maintains a substantial 20-25\% contribution. These results confirm our design principle: the global branch provides consistent balanced knowledge across all classes, while the personalized branch dynamically adjusts its contribution to compensate for data scarcity.

\section{Conclusion}

In this work, we explored the long-tail problem in personalized federated learning based on PEFT fine-tuning of foundation models. We observe that conventional approaches erode the pre-trained model’s inherent class balance and propagate bias to client-specific adaptations. We introduced FedPuReL, a framework that maintains global balance via zero-shot-guided gradient purification and supports unbiased personalization through residual learning in the output space. Extensive experiments on diverse datasets and imbalance factors showed that FedPuReL outperforms baselines for both global and personalized models, even without class priors. These results highlight how pre-trained knowledge can alleviate class imbalances in federated settings, fostering more equitable collaborative learning.

\section*{Acknowledgements}
This study was supported in part by the National Natural Science Foundation of China under Grants 62431004, 62376233, 62502402 and 62476063; in part by the Natural Science Foundation of Fujian Province under Grant 2024J09001; in part by the Natural Science Foundation of Guangdong Province under Grant 2025A1515011293; and in part by Xiaomi Young Talents Program.

{
    \small
    \bibliographystyle{ieeenat_fullname}
    \bibliography{main}
}

\clearpage
\setcounter{page}{1}
\maketitlesupplementary
\appendix

This appendix provides further details, results, and analyses that could not be included in the main paper owing to space constraints. The content is organized as follows:
\begin{itemize}
    \item Sec.~\ref{sec:notation} provides the notation table for FedPuReL.
    \item Sec.~\ref{APP:Algorithm} presents the complete algorithm description.
    \item Sec.~\ref{app:additional_analysis} provides additional analysis of FedPuReL, including complementary branch contributions and gradient alignment dynamics.
    \item Sec.~\ref{app:Additional Experiments} presents further experimental results, including robustness to data heterogeneity, hyperparameter studies, and convergence analysis.
    \item Sec.~\ref{app:discussion} discusses the key insights and implications of FedPuReL.
\end{itemize}

\section{Notation Table}
\label{sec:notation}
We provide the notation table in Tab.~\ref{tab:notation}.

\begin{table}[ht]
\caption{Notation table for FedPuReL.}
\label{tab:notation}
\centering
\footnotesize
\resizebox{0.48\textwidth}{!}{
\setlength\tabcolsep{2.5pt}
\renewcommand\arraystretch{1.15}
\begin{tabular}{cl}
\hline\thickhline
\rowcolor{gray!20}
\textbf{Notation} & \textbf{Description} \\
\hline\hline
\multicolumn{2}{l}{\textcolor{gray!60}{\textit{Federated Learning Setup}}} \\
\cellcolor{gray!10}$K$ & \cellcolor{gray!10}Number of clients \\
$\mathcal{D}_k$ & Local dataset of client $k$ \\
\cellcolor{gray!10}$n_k$ & \cellcolor{gray!10}Number of samples in client $k$ \\
$\mathcal{S}_t$ & Set of selected clients at round $t$ \\
\cellcolor{gray!10}$C$ & \cellcolor{gray!10}Number of classes \\
$\mathrm{IF}$ & Imbalance factor ($n_1 / n_C$) \\
\cellcolor{gray!10}$\alpha_{\text{dir}}$ & \cellcolor{gray!10}Dirichlet parameter for heterogeneity \\
\hdashline
\multicolumn{2}{l}{\textcolor{gray!60}{\textit{Model Parameters}}} \\
\cellcolor{gray!10}$\boldsymbol{W}$ & \cellcolor{gray!10}Frozen CLIP backbone parameters \\
$\boldsymbol{\phi}_g$ & Global shared PEFT parameters \\
\cellcolor{gray!10}$\boldsymbol{\phi}_k$ & \cellcolor{gray!10}Personalized PEFT parameters for client $k$ \\
$f_{\text{img}}(\cdot)$, $f_{\text{text}}(\cdot)$ & Image and text encoders \\
\hdashline
\multicolumn{2}{l}{\textcolor{gray!60}{\textit{Predictions and Logits}}} \\
\cellcolor{gray!10}$\mathbf{x}$, $y$ & \cellcolor{gray!10}Input image and label \\
$\mathbf{z}(\mathbf{x})$ & Zero-shot logits \\
\cellcolor{gray!10}$\mathbf{f}(\mathbf{x})$ & \cellcolor{gray!10}Fine-tuned logits \\
$\mathbf{l}_{\text{G}}(\mathbf{x})$ & Global branch logits \\
\cellcolor{gray!10}$\mathbf{l}_{\text{P}}^k(\mathbf{x})$ & \cellcolor{gray!10}Personalized branch logits for client $k$ \\
\hdashline
\multicolumn{2}{l}{\textcolor{gray!60}{\textit{Metrics and Temperature}}} \\
\cellcolor{gray!10}$\tau$, $\tau_{zs}$, $\tau_{ft}$ & \cellcolor{gray!10}Temperature parameters \\
$D_{\text{TKL}}(\cdot \| \cdot)$ & Temperature-aligned KL divergence \\
\cellcolor{gray!10}$\beta(V)$ & \cellcolor{gray!10}Balancedness metric \\
$H(\cdot)$ & Entropy function \\
\hdashline
\multicolumn{2}{l}{\textcolor{gray!60}{\textit{Gradients and Losses}}} \\
\cellcolor{gray!10}$\mathbf{g}_{\text{task}}$ & \cellcolor{gray!10}Task gradient from cross-entropy loss \\
$\mathbf{g}_{\text{align}}$ & Alignment gradient from TKL divergence \\
\cellcolor{gray!10}$\tilde{\mathbf{g}}_{\text{task}}$ & \cellcolor{gray!10}Purified task gradient \\
$\mathcal{L}_{\text{align}}$ & Alignment loss \\
\cellcolor{gray!10}$\mathcal{L}_{\text{fusion}}^k$ & \cellcolor{gray!10}Fusion loss for additive combination \\
$\mathcal{L}_{\text{personal}}^k$ & Personalization loss \\
\cellcolor{gray!10}$\lambda$ & \cellcolor{gray!10}Weight balancing fusion and personal losses \\
\hline\thickhline
\end{tabular}
}
\end{table}

\section{Algorithm}
\label{APP:Algorithm}
We provide the algorithm description in Algorithm~\ref{alg:fedpurel}.

\begin{algorithm}[ht]
\caption{FedPuReL}
\label{alg:fedpurel}
\begin{algorithmic}[1]
\Require Number of clients $K$, global rounds $T$, personalization rounds $T_p$, balancedness weight $\lambda$, frozen CLIP weights $\boldsymbol{W}$.
\Ensure Global PEFT parameters $\boldsymbol{\phi}_g$, personalized parameters $\{\boldsymbol{\phi}_k\}_{k=1}^K$.

\State \textcolor{adaptive_blue}{\textbf{// Phase 1: Global Balanced Training}}
\State Initialize $\boldsymbol{\phi}_g^{(0)}$
\For{$t = 1, \ldots, T$}
  \State Select active clients $\mathcal{S}_t \subseteq [K]$
  \ForAll{client $k \in \mathcal{S}_t$ \textbf{in parallel}}
    \State Receive $\boldsymbol{\phi}_g^{(t)}$ from server
      \State Compute $\mathbf{z}(\mathbf{x})$, $\mathbf{f}(\mathbf{x}; \boldsymbol{W}, \boldsymbol{\phi}_g)$
      \State $\mathbf{g}_{\text{align}} \gets \nabla_{\boldsymbol{\phi}_g} D_{\text{TKL}}(\sigma_{\tau_{zs}}(\mathbf{z}) \| \sigma_{\tau_{ft}}(\mathbf{f}))$
      \State $\mathbf{g}_{\text{task}} \gets \nabla_{\boldsymbol{\phi}_g} \text{CE}(y, \sigma(\mathbf{f}))$
      \State \textbf{// Gradient Purification}
      \If{$\langle \mathbf{g}_{\text{task}}, \mathbf{g}_{\text{align}} \rangle < 0$}
        \State $\tilde{\mathbf{g}}_{\text{task}} \gets \mathbf{g}_{\text{task}} - \frac{\langle \mathbf{g}_{\text{task}}, \mathbf{g}_{\text{align}} \rangle}{\|\mathbf{g}_{\text{align}}\|^2} \mathbf{g}_{\text{align}}$
      \Else
        \State $\tilde{\mathbf{g}}_{\text{task}} \gets \mathbf{g}_{\text{task}}$
      \EndIf
      \State Update $\boldsymbol{\phi}_g$ with $\tilde{\mathbf{g}}_{\text{task}}$
    \State Upload $\boldsymbol{\phi}_g^{k,(t)}$ to server
  \EndFor
  \State $\boldsymbol{\phi}_g^{(t+1)} \gets \sum_{k \in \mathcal{S}_t} \frac{n_k}{\sum_{j \in \mathcal{S}_t} n_j} \boldsymbol{\phi}_g^{k,(t)}$
\EndFor

\State \textcolor{adaptive_blue}{\textbf{// Phase 2: Personalized Residual Learning}}
\State Freeze $\boldsymbol{\phi}_g^{(T)}$
\ForAll{client $k \in [K]$ \textbf{in parallel}}
  \State Initialize $\boldsymbol{\phi}_k$
  \For{$t_p = 1, \ldots, T_p$}
      \State $\mathbf{l}_{\text{G}} \gets f(\mathbf{x}; \boldsymbol{W}, \boldsymbol{\phi}_g^{(T)})$, $\mathbf{l}_{\text{P}}^k \gets f(\mathbf{x}; \boldsymbol{W}, \boldsymbol{\phi}_k)$
      \State $\mathcal{L}^k \gets (1-\lambda) \text{CE}(y, \sigma(\mathbf{l}_{\text{G}} + \mathbf{l}_{\text{P}}^k)) + \lambda \text{CE}(y, \sigma(\mathbf{l}_{\text{P}}^k))$
      \State Update $\boldsymbol{\phi}_k$ with $\nabla_{\boldsymbol{\phi}_k} \mathcal{L}^k$
    \EndFor
\EndFor
\State \Return $(\boldsymbol{W}, \boldsymbol{\phi}_g^{(T)})$, $\{(\boldsymbol{W}, \boldsymbol{\phi}_g^{(T)}, \boldsymbol{\phi}_k)\}_{k=1}^K$
\end{algorithmic}
\end{algorithm}

\section{Additional Analysis of FedPuReL}\label{app:additional_analysis}

\noindent \textbf{Analysis 4: Global and personalized branches exhibit complementary contributions.} 
Our design separates balanced global knowledge from personalized local adaptation, which should lead to different contribution patterns across class types. Specifically, head classes with abundant samples should rely more on global knowledge, while tail classes with scarce data should require stronger personalized corrections. 

To evaluate this, we decompose each correct prediction to identify the dominant contributor between the global and personalized branches. For each correctly classified sample $x$ with ground-truth label $y$, we compare the logits from both branches: $l_G^{(y)}(x)$ from the global branch and $l_P^{(y)}(x)$ from the personalized branch. We attribute the prediction to the branch with the higher logit value for the correct class, i.e., the prediction is attributed to the global branch if $l_G^{(y)}(x) > l_P^{(y)}(x)$, and to the personalized branch otherwise. We then compute the percentage of correct predictions attributed to each branch across all classes, grouped by their sample frequency.

As illustrated in Fig.~\ref{fig:any2}, as class frequency decreases from head to tail, the personalized branch contribution increases, while the global branch contribution decreases correspondingly. Notably, even for tail classes where personalization dominates, the global branch maintains a substantial 20-25\% contribution. These results confirm our design principle: the global branch provides consistent balanced knowledge across all classes, while the personalized branch dynamically adjusts its contribution to compensate for data scarcity.

\begin{figure}[ht]
\centering
\includegraphics[width=0.7\linewidth]{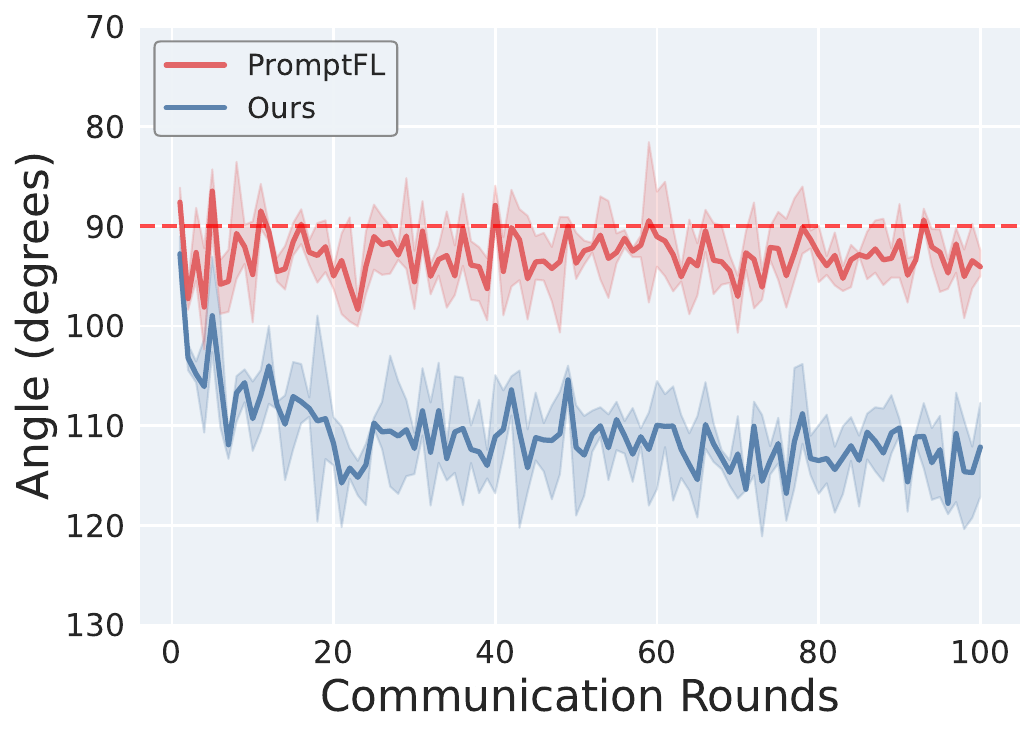}
\caption{Angle between $\mathbf{g}_{\text{task}}$ and $\mathbf{g}_{\text{align}}$ during training on CIFAR-100-LT.}

\label{fig:gradient_angle}
\end{figure}

\noindent \textbf{Analysis 5: Gradient Alignment Dynamics.}
Fig.~\ref{fig:gradient_angle} visualizes the angle between task gradient $\mathbf{g}_{\text{task}}$ and alignment gradient $\mathbf{g}_{\text{align}}$ during training. Baseline methods exhibit convergence to approximately 90° (orthogonal), reflecting the well-known property that high-dimensional random vectors tend to be mutually orthogonal~\cite{cai2013distributions}. This orthogonality explains balance degradation, as task optimization proceeds without consideration of zero-shot alignment. In contrast, FedPuReL consistently maintains obtuse angles (\>90°), revealing that the original task gradient actively conflicts with balance preservation and would otherwise compromise zero-shot knowledge. Our purification operation detects these conflicts and removes incompatible components, ensuring updates remain aligned with balanced representations.


\section{Additional Experiments}
\label{app:Additional Experiments}

\begin{figure}[ht]
\centering
\includegraphics[width=\linewidth]{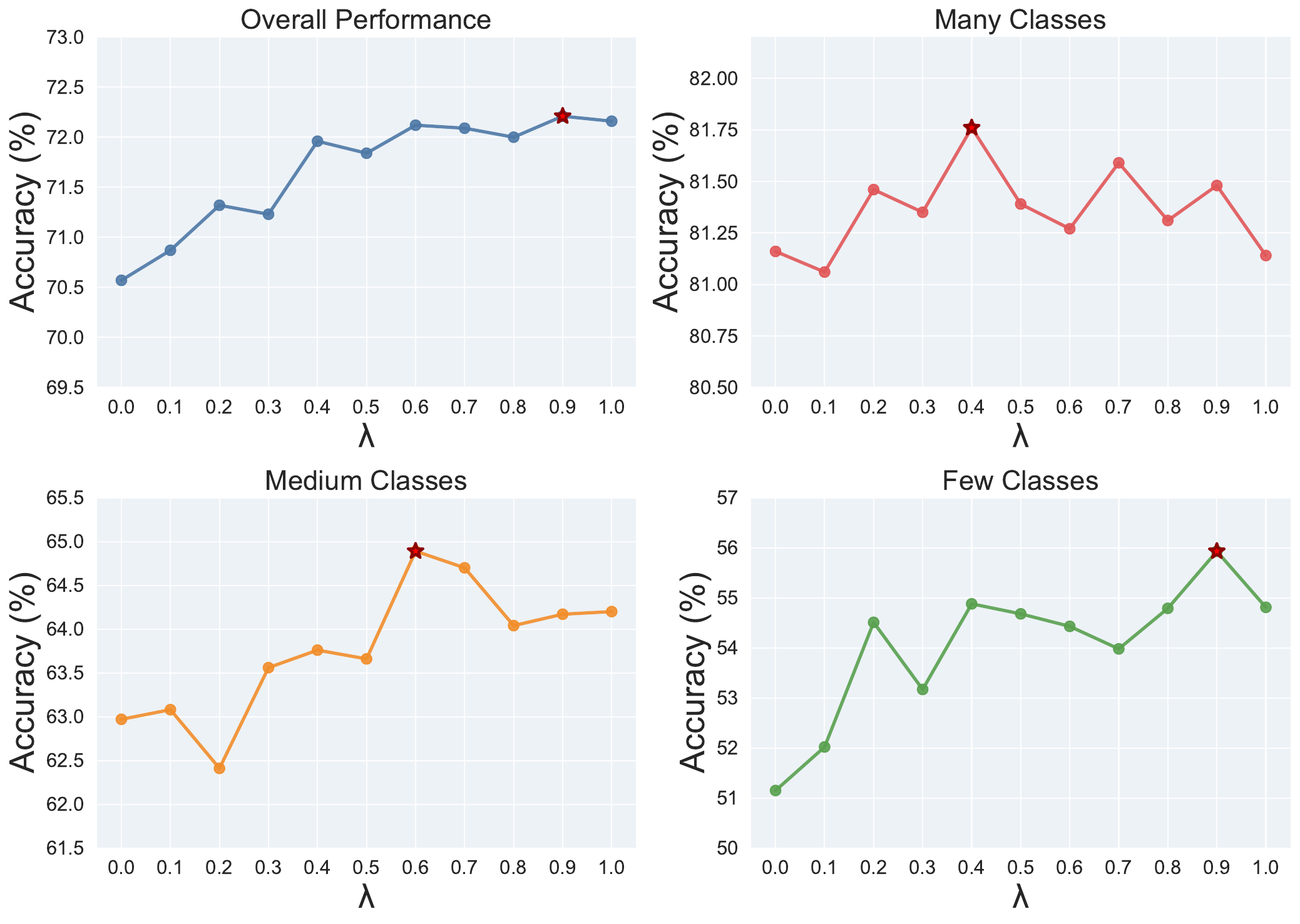}
\caption{\textbf{Ablation study on $\lambda$ in Eq.~\ref{fusion loss}.} Performance across overall, Head, Mid, and Tail classes on CIFAR-100-LT with varying $\lambda$ values. Red stars mark optimal points.}
\label{fig:ablation_lambda}
\end{figure}

\begin{figure}[!t]
\centering
\begin{subfigure}[b]{\linewidth}
    \centering
    \includegraphics[width=\linewidth]{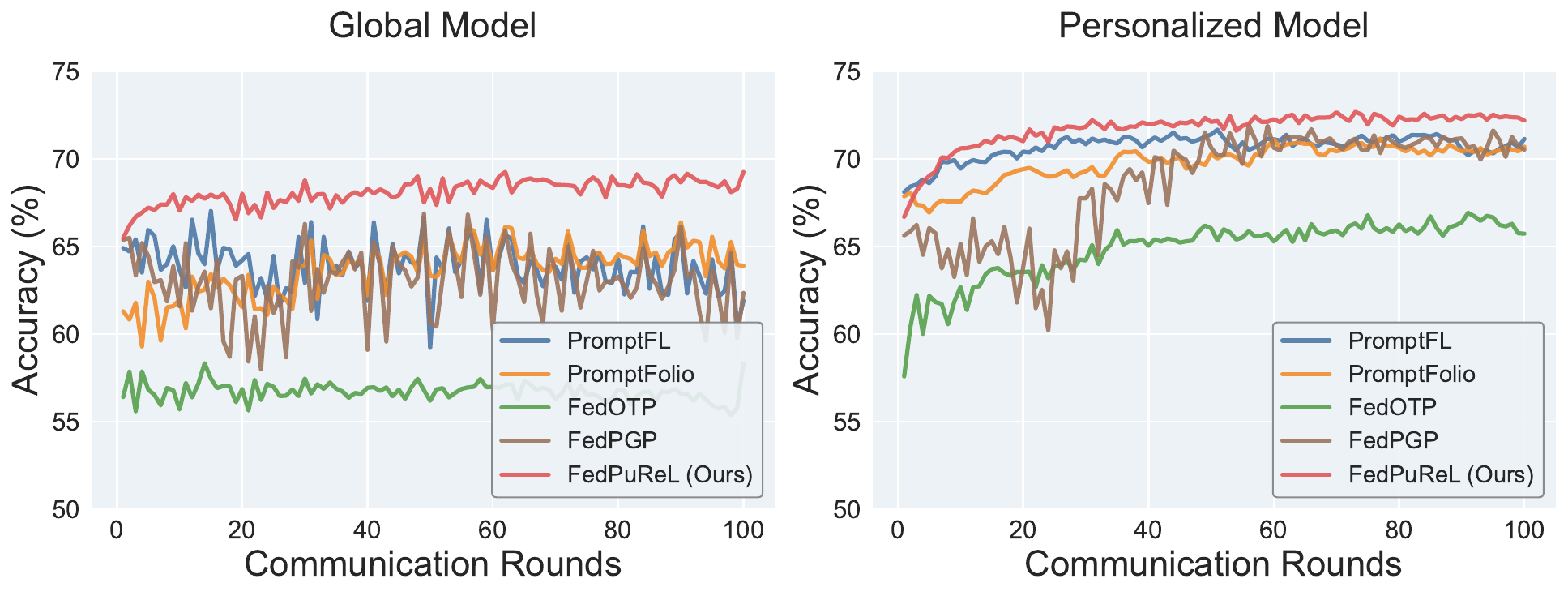}
    \caption*{(a) Overall Accuracy}
\end{subfigure}
\hfill
\begin{subfigure}[b]{\linewidth}
    \centering
    \includegraphics[width=\linewidth]{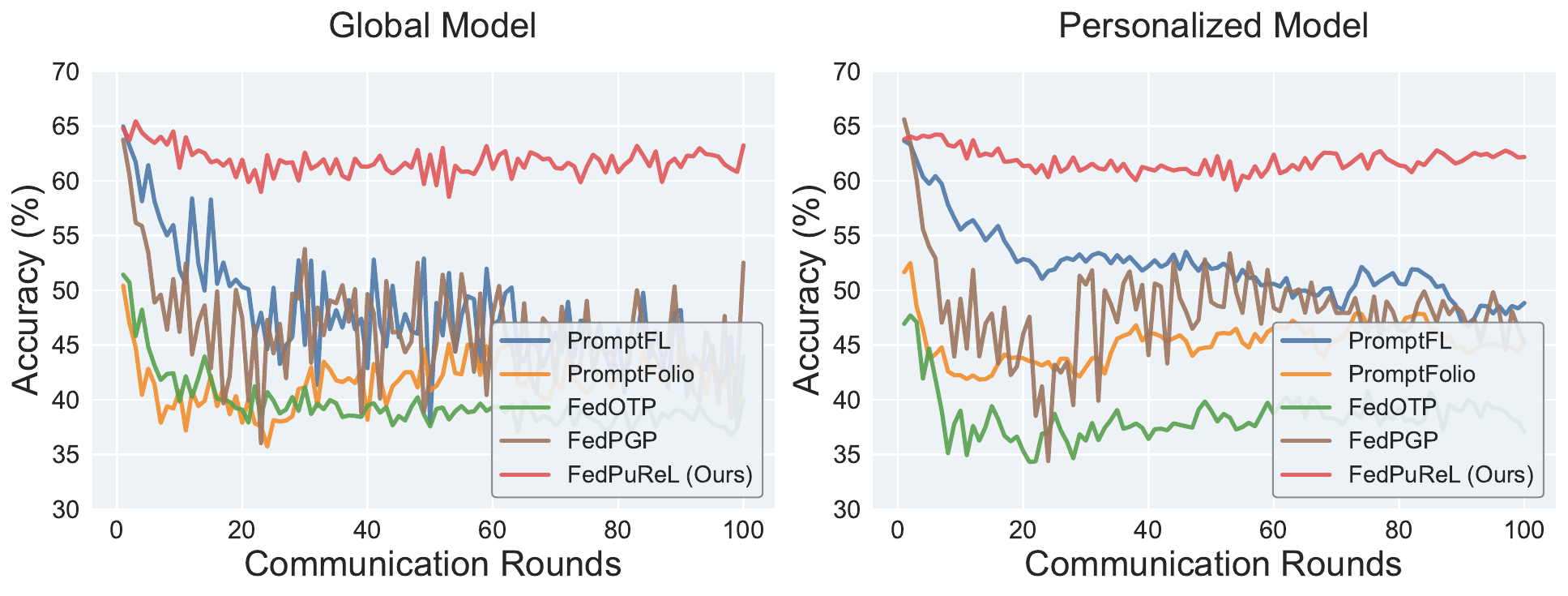}
    \caption*{(b) Few Class Accuracy}
\end{subfigure}
\vspace{-0.5em}
\caption{\textbf{Convergence of average accuracy} compared to Prompt-based SOTA method on CIAFR-100-LT}
\label{fig:acc_trend}
\end{figure}

\noindent \textbf{Ablation of TKL vs.\ Standard KL.}
Fig.~6 in the main paper qualitatively demonstrates that TKL distinguishes balanced adaptation from biased divergence, whereas standard KL exhibits large values even on balanced data due to confidence shifts.
To further validate this quantitatively, we replace TKL with standard KL divergence in the gradient purification module and evaluate on CIFAR-100-LT.
As shown in Table~\ref{tab:tkl_ablation}, TKL consistently outperforms standard KL in both accuracy and balancedness across global and personalized settings.
This confirms that temperature alignment is essential for isolating structural distributional shift from benign confidence growth during fine-tuning, enabling more effective gradient purification.

\begin{table}[ht]
\caption{\textbf{Ablation of TKL vs.\ standard KL} on CIFAR-100-LT.}
\label{tab:tkl_ablation}
\centering
\footnotesize
\setlength\tabcolsep{4pt}
\renewcommand\arraystretch{1.15}
\begin{tabular}{r||cc|cc}
\hline\thickhline
\rowcolor{gray!20}
 & \multicolumn{2}{c|}{\textbf{Accuracy (\%)}} & \multicolumn{2}{c}{\textbf{Balancedness (\%)}} \\
\cline{2-5}
\rowcolor{gray!20}
\textbf{Metric} & \textbf{GM} & \textbf{PM} & \textbf{GM} & \textbf{PM} \\
\hline\hline
\cellcolor{gray!10} KL & \cellcolor{gray!10} 67.92 & \cellcolor{gray!10} 71.62 & \cellcolor{gray!10} 25.03 & \cellcolor{gray!10} 27.84 \\
\cellcolor[HTML]{D7F6FF}\textbf{TKL (ours)} & \cellcolor[HTML]{D7F6FF}\textbf{69.77} & \cellcolor[HTML]{D7F6FF}\textbf{73.37} & \cellcolor[HTML]{D7F6FF}\textbf{27.87} & \cellcolor[HTML]{D7F6FF}\textbf{30.02} \\
\hline\thickhline
\end{tabular}
\vspace{-8pt}
\end{table}

\noindent \textbf{Robustness to Data Heterogeneity.} We evaluate FedPuReL's robustness across varying degrees of client heterogeneity by adjusting the Dirichlet parameter $\alpha \in \{0.1, 0.5, 1, 5\}$, where smaller $\alpha$ indicates higher non-IID heterogeneity. As shown in Table~\ref{tab:cifar100_lt_alpha}, FedPuReL consistently outperforms all baselines across different heterogeneity levels in both global and personalized settings. This robustness stems from our gradient purification mechanism, which anchors local updates to the zero-shot foundation model, effectively preventing divergence caused by heterogeneous local distributions. The consistent performance across all $\alpha$ values demonstrates that FedPuReL successfully preserves balanced knowledge while adapting to diverse client-specific distributions.

\noindent \textbf{Hyperparameter Study.} We analyze the impact of the balancing coefficient $\lambda$ in Eq.~\ref{fusion loss}, which controls the trade-off between fusion loss $\mathcal{L}^k_{\text{fusion}}$ and personalization loss $\mathcal{L}^k_{\text{personal}}$. As shown in Fig.~\ref{fig:ablation_lambda}, we observe that $\lambda$ exhibits different optimal values across class groups: $\lambda=0.9$ achieves the best overall and tail class performance, while moderate values favor head and mid classes. This demonstrates that higher $\lambda$ values, which emphasize the personalized branch, are particularly beneficial for tail classes that require stronger client-specific corrections. Notably, head classes maintain relatively stable performance across different $\lambda$ values, benefiting from the robust global branch. We set $\lambda=0.9$ as the default in our experiments to balance overall accuracy with tail class performance.

\noindent \textbf{Model Convergence Analysis.} Fig.~\ref{fig:acc_trend} visualizes the training dynamics of FedPuReL compared to state-of-the-art prompt-based methods on CIFAR-100-LT. For overall accuracy (Fig.~\ref{fig:acc_trend}a), FedPuReL maintains stable performance throughout training in both global and personalized settings. In contrast, baseline methods exhibit significant fluctuations and fail to achieve comparable performance. More critically, for few-shot class accuracy (Fig.~\ref{fig:acc_trend}b), FedPuReL maintains consistently superior performance on tail classes, while baselines show severe degradation or instability. This stability stems from our gradient purification mechanism, which prevents the model from drifting away from the balanced zero-shot anchor during adaptation. The consistent gap between FedPuReL and baselines throughout training validates that preserving balanced knowledge is essential for sustained performance on long-tailed federated data.

\begin{table}[ht]\footnotesize
\caption{\textbf{Comparison with state-of-the-art methods} on CIFAR-100-LT under different degrees of data heterogeneity ($\alpha$) for global models (GM) and personalized models (PM).}
\centering
\scriptsize{
\resizebox{\linewidth}{!}{
\setlength\tabcolsep{1.5pt}
\renewcommand\arraystretch{1.1}
\begin{tabular}{r||cc|cc|cc|cc}
\cline{2-9}
\hline\thickhline
\rowcolor{gray!20}
\textbf{CIFAR-100-LT} &
  \multicolumn{2}{c|}{\textbf{$\alpha = 0.1$}} &
  \multicolumn{2}{c|}{\textbf{$\alpha = 0.5$}} &
  \multicolumn{2}{c|}{\textbf{$\alpha = 1$}} &
  \multicolumn{2}{c}{\textbf{$\alpha = 5$}} \\
\cline{2-9}
\rowcolor{gray!20}
\textbf{Method} & \textbf{GM} & \textbf{PM} & \textbf{GM} & \textbf{PM} & \textbf{GM} & \textbf{PM} & \textbf{GM} & \textbf{PM} \\
\hline\hline
Zero-shot & 64.82 & 64.79 & 64.82 & 64.79 & 64.82 & 64.79 & 64.82 & 64.79 \\
\hdashline
\multicolumn{9}{l}{\textcolor{gray!60}{\textit{Prompt-based}}} \\
\cellcolor{gray!10}PromptFL~\cite{guo2023promptfl}
 & \cellcolor{gray!10}63.72 & \cellcolor{gray!10}72.18 & \cellcolor{gray!10}65.25 & \cellcolor{gray!10}71.97 & \cellcolor{gray!10}61.98 & \cellcolor{gray!10}70.52 & \cellcolor{gray!10}64.95 & \cellcolor{gray!10}71.38 \\
PromptFolio~\cite{pan2024promptfolio}
 & 64.32 & 72.83 & 64.79 & 72.19 & 64.10 & 70.54 & 65.68 & 71.10 \\
\cellcolor{gray!10}FedOTP~\cite{li2024fedotp}
 & \cellcolor{gray!10}57.27 & \cellcolor{gray!10}70.87 & \cellcolor{gray!10}55.75 & \cellcolor{gray!10}68.28 & \cellcolor{gray!10}58.02 & \cellcolor{gray!10}65.23 & \cellcolor{gray!10}54.88 & \cellcolor{gray!10}65.77 \\
FedPGP~\cite{cui2024fedpgp}
 & 61.66 & 73.10 & 62.12 & 72.56 & 62.96 & 67.83 & 63.41 & 70.41 \\
 PromptFolio+Fed-GraB~\cite{Fed-Grab}
 & \underline{64.97} & \underline{75.13} & \underline{65.43} & \underline{73.26} & \underline{65.14} & \underline{71.79} & \underline{66.23} & \underline{72.18} \\
 
\cellcolor[HTML]{D7F6FF}\textbf{FedPuReL (ours)}
 & \cellcolor[HTML]{D7F6FF}\textbf{68.88} & \cellcolor[HTML]{D7F6FF}\textbf{77.72} & \cellcolor[HTML]{D7F6FF}\textbf{69.05} & \cellcolor[HTML]{D7F6FF}\textbf{74.97} & \cellcolor[HTML]{D7F6FF}\textbf{69.77} & \cellcolor[HTML]{D7F6FF}\textbf{73.37} & \cellcolor[HTML]{D7F6FF}\textbf{70.54} & \cellcolor[HTML]{D7F6FF}\textbf{73.71} \\
\textcolor{gray!60}&
\blueup{3.91} & \blueup{2.59} &
\blueup{3.62} & \blueup{1.71} &
\blueup{4.63} & \blueup{1.58} &
\blueup{4.31} & \blueup{1.53} \\
\hline
\multicolumn{9}{l}{\textcolor{gray!60}{\textit{LoRA-based}}} \\
CLIPLoRA~\cite{cliplora}
 & 74.83 & 78.53 & 75.50 & 78.75 & 75.02 & 77.80 & 76.12 & 79.12 \\
FedSA-LoRA~\cite{wang2024fedsalora}
 & 73.53 & \underline{80.47} & 73.41 & \underline{79.16} & 74.48 & \underline{81.68} & 73.03 & 76.91 \\
 CLIPLoRA+Fed-GraB~\cite{Fed-Grab}
 & \underline{75.30} & 79.12 & \underline{76.43} & 78.97 & \underline{76.35} & 77.91 & \underline{77.31} & \underline{79.62} \\
 
\cellcolor[HTML]{D7F6FF}\textbf{FedPuReL (ours)}
 & \cellcolor[HTML]{D7F6FF}\textbf{76.71} & \cellcolor[HTML]{D7F6FF}\textbf{81.76} & \cellcolor[HTML]{D7F6FF}\textbf{76.74} & \cellcolor[HTML]{D7F6FF}\textbf{80.29} & \cellcolor[HTML]{D7F6FF}\textbf{77.56} & \cellcolor[HTML]{D7F6FF}\textbf{84.62} & \cellcolor[HTML]{D7F6FF}\textbf{78.20} & \cellcolor[HTML]{D7F6FF}\textbf{80.27} \\
\textcolor{gray!60} &
\blueup{1.41} & \blueup{1.29} &
\blueup{0.31} & \blueup{1.13} &
\blueup{1.21} & \blueup{2.94} &
\blueup{0.89} & \blueup{0.65} \\

\hline
\multicolumn{9}{l}{\textcolor{gray!60}{\textit{Adapter-based}}} \\
FedClip~\cite{lu2023fedclip}
 & 63.92 & 68.02 & 64.54 & 68.64 & 64.09 & 66.87 & 64.35 & 68.19 \\

 FedClip+Fed-GraB~\cite{Fed-Grab}
 & \underline{64.69} & \underline{68.18} & \underline{65.72} & \underline{68.98} & \underline{65.40} & \underline{67.20} & \underline{66.02} & \underline{68.49} \\
 
\cellcolor[HTML]{D7F6FF}\textbf{FedPuReL (ours)}
 & \cellcolor[HTML]{D7F6FF}\textbf{66.58} & \cellcolor[HTML]{D7F6FF}\textbf{68.98} & \cellcolor[HTML]{D7F6FF}\textbf{66.83} & \cellcolor[HTML]{D7F6FF}\textbf{69.38} & \cellcolor[HTML]{D7F6FF}\textbf{67.54} & \cellcolor[HTML]{D7F6FF}\textbf{68.42} & \cellcolor[HTML]{D7F6FF}\textbf{67.81} & \cellcolor[HTML]{D7F6FF}\textbf{69.05} \\
\textcolor{gray!60}&
\blueup{1.89} & \blueup{0.80} &
\blueup{1.11} & \blueup{0.40} &
\blueup{2.14} & \blueup{1.22} &
\blueup{1.79} & \blueup{0.56} \\

\end{tabular}}
}
\vspace{-12pt}
\label{tab:cifar100_lt_alpha}
\end{table}

\section{Discussion}\label{app:discussion}
FedPuReL preserves balanced knowledge in foundation models while enabling effective personalization. Gradient purification maintains balancedness close to zero-shot model throughout training (cf. Analysis~\hyperlink{analysis:1}{1} in Sec. 4.5). Residual learning then enables unbiased personalization through output-space offsets: Analysis~\hyperlink{analysis:4}{4} reveals complementary contributions where the global branch provides consistent knowledge across all classes while the personalized branch dynamically compensates for data scarcity in tail classes. Additionally, unlike methods requiring explicit class priors~\cite{zhang2024rucr,Fed-Grab}, our approach leverages implicit balanced priors in foundation models, eliminating the need for class distribution that pose privacy risks while exceeding prior-based performance (Sec.~\ref{Comparison with State-of-the-Art}). By reducing distributional divergence and client drift (Analysis~\hyperlink{analysis:2}{2}--\hyperlink{analysis:3}{3}), FedPuReL achieves superior convergence without explicit rebalancing.

\end{document}